\documentclass[10pt,twocolumn,letterpaper]{article}

\usepackage[pagenumbers]{cvpr} 

\usepackage{multirow}

\definecolor{cvprblue}{rgb}{0.21,0.49,0.74}
\usepackage{algorithm}
\usepackage{algorithmic}
\usepackage[pagebackref,breaklinks,colorlinks,allcolors=cvprblue]{hyperref}

\def\paperID{14460} 
\def\confName{CVPR}
\def\confYear{2026}

\title{MDG: Masked Denoising Generation for Multi-Agent \\ Behavior Modeling in Traffic Environments}

\author{Zhiyu Huang, Zewei Zhou, Tianhui Cai, Yun Zhang, Jiaqi Ma \\
University of California, Los Angeles\\
{\tt\small \{zhiyuh, zeweizhou, tianhui, yun666, jiaqima\}@ucla.edu}
}

\begin{document}
\maketitle

\begin{abstract}
Modeling realistic and interactive multi-agent behavior is critical to autonomous driving and traffic simulation. However, existing diffusion and autoregressive approaches are limited by iterative sampling, sequential decoding, or task-specific designs, which hinder efficiency and reuse. We propose \textbf{Masked Denoising Generation (MDG)}, a unified generative framework that reformulates multi-agent behavior modeling as the reconstruction of independently noised spatiotemporal tensors. Instead of relying on diffusion time steps or discrete tokenization, MDG applies continuous, per-agent and per-timestep noise masks that enable localized denoising and controllable trajectory generation in a single or few forward passes. This mask-driven formulation generalizes across open-loop prediction, closed-loop simulation, motion planning, and conditional generation within one model. Trained on large-scale real-world driving datasets, MDG achieves competitive closed-loop performance on the Waymo Sim Agents and nuPlan Planning benchmarks, while providing efficient, consistent, and controllable multi-agent trajectory generation. These results position MDG as a simple yet versatile paradigm for multi-agent behavior modeling.
\end{abstract}     
\section{Introduction}
\label{sec:intro}

Multi-agent behavior modeling is a cornerstone for enabling safe and interactive autonomous systems in complex real-world environments \cite{li2024multiagent}. Accurate and controllable behavior generation supports a range of downstream tasks, from open-loop motion prediction \cite{zhou2023qcnext, ruan2024learning, zhou2024smartpretrain} to traffic simulation \cite{zhao2025drope, rowe2024ctrl} and closed-loop planning \cite{cheng2024pluto, wu2022trajectory}. These tasks are typically addressed using distinct models and objectives: prediction emphasizes accuracy and diversity, simulation demands controllability and interactivity, and planning prioritizes consistency and efficiency. This separation prevents models from generalizing or being reused across tasks \cite{huang2024versatile, jiangscenediffuser, zheng2025diffusionbased}, hindering scalable autonomy development. 

\begin{figure}[t]
\centering
\includegraphics[width=\linewidth]{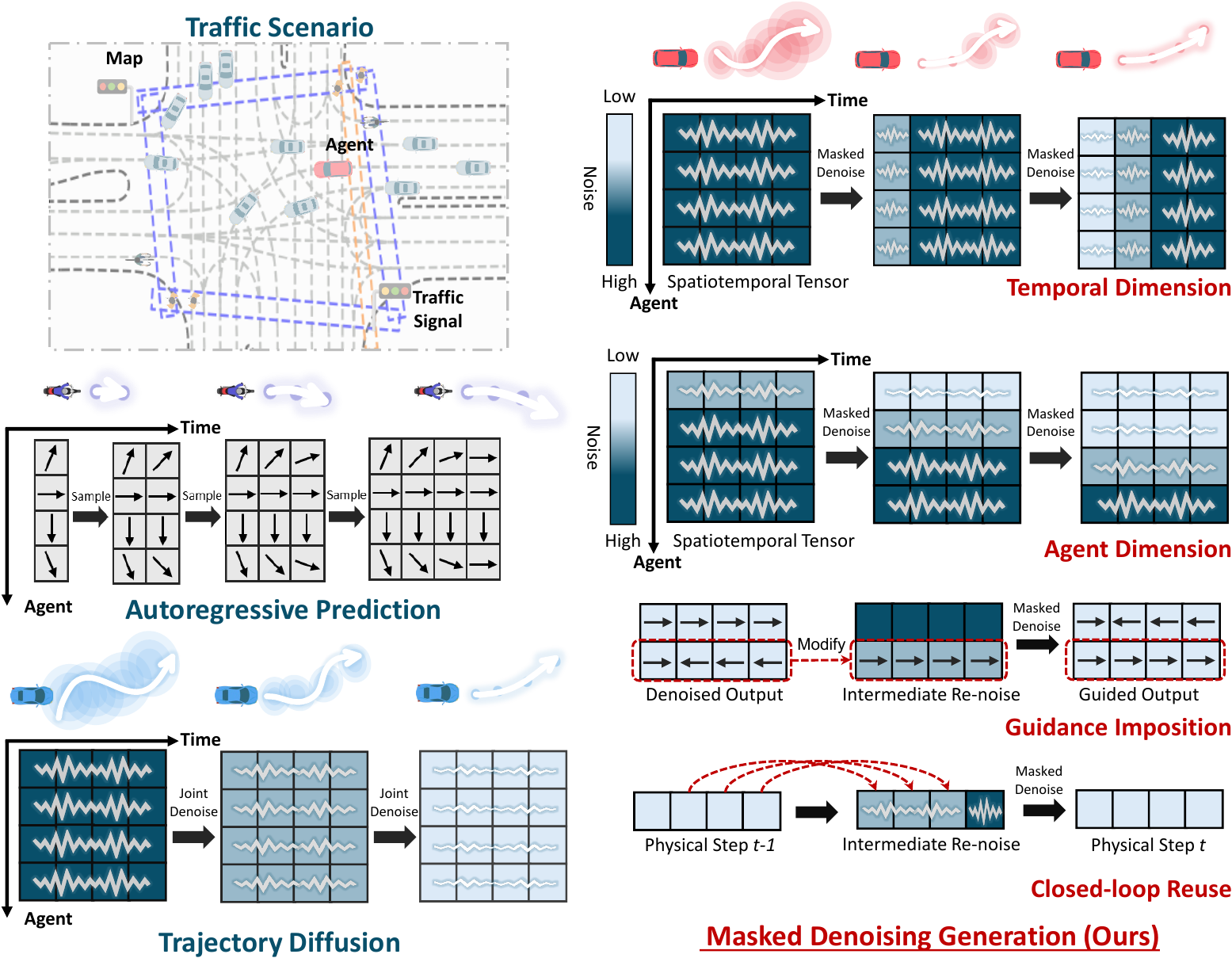}
\caption{
Comparison of MDG with existing trajectory generation paradigms. MDG denoises masked spatiotemporal tensors under varied noise-masking patterns, enabling temporal-wise, agent-wise generation, guided conditioning, and closed-loop reuse. Unlike autoregressive models, MDG predicts full-sequence multi-agent futures in a single step, and unlike joint trajectory diffusion, it supports fine-grained control with efficient, flexible sampling.}
\label{fig:hero}
\vspace{-0.3cm}
\end{figure}


Recent advances in generative modeling have improved realism and diversity by learning joint scene-level distributions over agent behaviors \cite{tan2021scenegen, zhao2024kigras, tan2024promptable}. Two major paradigms have emerged: diffusion-based \cite{zhong2023language, chang2025safe} and autoregressive (AR) approaches \cite{zhou2024behaviorgpt, wu2024smart, philion2024trajeglish}. However, both paradigms exhibit inherent limitations. Joint trajectory diffusion models \cite{huang2024versatile, zhong2023guided, jiang2023motiondiffuser} operate through iterative noise removal at the scene level, which limits controllability, increases computational cost when guidance is used, and may produce out-of-distribution actions as denoising steps accumulate \cite{huang2024versatile}. AR generative models, which discretize continuous states or actions into tokens and rely on sequential sampling, struggle to capture continuous dynamics and enforce long-term guidance for specific agents. Recent extensions such as Diffusion Forcing \cite{chen2024diffusion} and Masked Diffusion \cite{sahoo2024simple, rector2024steering} introduce per-element noise or partial masking, but still rely on diffusion time-stepping or categorical data. Consequently, existing approaches struggle to generate multi-agent behaviors that are efficient, consistent, and easily controllable across diverse tasks.

To address these limitations, we propose \textbf{Masked Denoising Generation (MDG)}, a continuous, structured, and noise mask-driven generative paradigm for multi-agent behavior modeling. MDG reformulates multi-agent trajectory generation as the reconstruction of independently noised spatiotemporal tensors rather than as iterative diffusion or token decoding. Each agent-time position receives an independent mask intensity that determines how much noise is applied, enabling the model to perform localized denoising and targeted conditioning within a single or a few forward passes. This design combines the controllability and efficiency of masked modeling with the expressiveness of continuous denoising, while avoiding the diffusion time axis \cite{sun2025noise}.
As illustrated in \cref{fig:hero}, MDG supports several inference modes, such as temporal or agent-wise denoising, conditional guidance, and closed-loop reuse. By introducing a continuous mask field over structured multi-agent trajectories, MDG provides a simple yet general framework to perform open-loop prediction, interactive simulation, and motion planning. The main contributions of this paper are summarized as follows:
\begin{enumerate}
\item We introduce Masked Denoising Generation (MDG), a generative paradigm that models multi-agent behaviors through mask-based denoising of spatiotemporal tensors, supporting diverse behavior modeling tasks.
\item We propose a per-agent, per-timestep mask field that regulates localized denoising and enables flexible inference modes, including temporal-wise, agent-wise, condition-guided generation, and closed-loop reuse.
\item We demonstrate that MDG is a general framework for behavior modeling, achieving competitive closed-loop results on Waymo Sim Agents and nuPlan benchmark, while supporting efficient and controllable generation.
\end{enumerate}

\section{Related Work}
\textbf{Multi-agent Behavior Modeling in Traffic Scenarios.}
Modeling the joint behaviors of multiple interacting agents is a central challenge in autonomous driving \cite{huang2022multi, baniodeh2025scaling, jia2022multi, rowe2023fjmp, jia2023hdgt, hu2025solving, liu2024multi}. Traditional predictive methods, such as MTR \cite{shi2024mtr++} and GameFormer \cite{huang2023gameformer}, directly decode future trajectories from historical states, achieving accurate but often marginalized predictions that fail to capture inter-agent dependencies. As generative modeling advances, recent approaches have shifted toward learning scene-level distributions over all agents’ futures, enabling richer and more interactive behaviors.
Models such as MotionLM \cite{seff2023motionlm} and MotionDiffuser \cite{jiang2023motiondiffuser} improve behavioral diversity but remain limited to pairwise or partially joint dependencies. More recent generative frameworks, including autoregressive Transformers (\eg, Trajeglish \cite{philion2024trajeglish}, BehaviorGPT \cite{zhou2024behaviorgpt}, SMART \cite{wu2024smart}) and diffusion-based models (\eg, VBD \cite{huang2024versatile}, SceneDiffuser \cite{jiangscenediffuser}), have advanced multi-agent prediction and simulation. Advanced training methods, such as closed-loop fine-tuning \cite{zhang2024closed} and reinforcement fine-tuning \cite{huang2024gen, pei2025advancing} have further enhanced performance. Despite this progress, both AR and diffusion paradigms have intrinsic limitations. Autoregressive models rely on discrete sequential decoding, which hinders long-horizon guidance and temporal coherence, while diffusion-based models require slow iterative sampling and struggle with fine-grained control. 
MDG unifies these paradigms by replacing sequential or iterative generation with a continuous, mask-conditioned denoising process that reconstructs the spatiotemporal tensor in a single step. This formulation enables MDG to produce consistent, diverse, and controllable multi-agent futures.

\noindent \textbf{Trajectory Diffusion Models.}
Diffusion models \cite{ho2020denoising, song2020denoising, li2024immiscible} have demonstrated strong performance in generating continuous and temporally consistent behaviors \cite{urain2024deep, zhou2025decoupled}. In autonomous driving, diffusion models have been applied to trajectory prediction (\eg, MotionDiffuser \cite{jiangscenediffuser}), scene simulation (\eg, SceneDiffuser \cite{jiangscenediffuser}, SceneDM \cite{guo2023scenedm}, VBD \cite{huang2024versatile}), and planning (\eg, Gen-Drive \cite{huang2024gen}, Diffusion-ES \cite{yang2024diffusion}, Diffusion-Planner \cite{zheng2025diffusionbased}).
However, existing approaches typically apply uniform noise across entire multi-agent sequences and rely on iterative denoising or integration of ordinary differential equations for sample generation, which increases computational cost and yields limited benefits under strong conditioning tasks (\eg, behavior generation in a traffic context). Recent flow-based variants for trajectory generation, such as Leapfrog \cite{mao2023leapfrog}, TrajFlow \cite{yan2025trajflow}, and MoFlow \cite{fu2025moflow}, still depend on such iterative processes. In contrast, MDG performs single-step masked reconstruction with per-timestep, per-agent noise, enabling localized conditioning and efficient generation while maintaining the expressive capacity.

\noindent \textbf{Masked Generative Modeling.}
Masked generative modeling has recently shown strong scalability across modalities such as language \cite{sahoo2024simple, rector2024steering}, video \cite{liu2024mardini}, and images \cite{gao2023masked}. These approaches train models to reconstruct masked inputs, offering an efficient alternative to stepwise generation. Inspired by this, MDG adopts a masked denoising formulation, introducing independent per-token noise levels for localized reconstruction and adaptive conditioning.
Masked modeling has been explored in trajectory prediction and generation (\eg, masked trajectory model \cite{wu2023masked} and Forecast-MAE \cite{cheng2023forecast}), which demonstrate that randomized masking improves robustness and inference flexibility. MDG generalizes these ideas by introducing a continuous mask field over structured spatiotemporal tensors, combining the efficiency of masked modeling with the expressive power of denoising.
\section{Masked Denoising Generation}

\subsection{Preliminary}

\textbf{Diffusion Models.}
Diffusion models \cite{ho2020denoising} are a family of generative models that learn to recover structured data from noise through iterative denoising. Let $q(\mathbf{x})$ represent the data distribution, and $\mathbf{x} \sim q(\mathbf{x})$ denote clean data sampled from this distribution. The forward diffusion process gradually adds Gaussian noise over $K$ steps:
\begin{equation}
\mathbf{z}_k = \sqrt{\bar \alpha_k} \mathbf{x} + \sqrt{1-\bar \alpha_k} \epsilon, \ \epsilon \sim \mathcal{N}(0, \mathbf{I}), 
\end{equation} 
where $\bar \alpha_k$ is a noise schedule controlling the magnitude of noise at step $k$. The reverse process learns to sequentially denoise $\mathbf{z}_k$ to recover the original data. In our context, $\mathbf{x}$ is structured as a spatiotemporal tensor $x_a^t$ with temporal axis $t$ and agent axis $a$.



\noindent\textbf{Diffusion Forcing.}
Diffusion forcing~\cite{chen2024diffusion} extends standard diffusion by assigning an independent noise level to each element in a sequence rather than a single global timestep.
This allows the model to treat some elements as nearly clean while others remain highly corrupted, effectively interpolating between next-token prediction and full-sequence diffusion. However, it still relies on multi-step denoising and is primarily applied to one-dimensional token sequences (\eg, text or video tokens), without modeling of structured spatiotemporal data.

\noindent \textbf{Masked Discrete Diffusion}.
Masked diffusion \cite{sahoo2024simple, rector2024steering} represents another form of partial corruption, typically used in discrete domains. These methods replace missing elements with a \texttt{[MASK]} token and train a model to reconstruct them from the visible context. While computationally efficient, they operate on categorical data and cannot handle continuous-valued signals or partially noised inputs.

\subsection{Masked Denoising}
We propose Masked Denoising Generation (MDG), a continuous spatiotemporal generalization of masked and diffusion-based paradigms. The key idea is to represent noise as a \emph{continuous mask field} applied independently to each agent-time position in the trajectory tensor. This allows single-step or few-step denoising, localized conditioning, and controllable scene generation.
\begin{itemize}
    \item Compared to standard diffusion, MDG supports single-step reconstruction for efficient prediction and also admits controlled iterative refinement when required.
    \item Compared to diffusion forcing, MDG generalizes per-element noise to continuous, structured spatiotemporal tensors with per-agent and per-timestep masking, enabling more flexible control.
    \item Compared to masked discrete diffusion, MDG represents masking as continuous Gaussian corruption (soft masks) instead of discrete mask tokens, permitting partial corruption and continuous outputs.
\end{itemize}

\noindent \textbf{Mask-driven Corruption Process.}
Each element of the trajectory tensor is assigned a noise level (mask intensity) indicating how strongly it should be perturbed. Let $\mathbf{m} \in [0,K]^{T \times N}$ ($N$ agents, $T$ timesteps) be a noise-level mask, where $m_a^t$ specifies the noise magnitude applied to position $x_a^t$. The forward-noising process is defined as:
\begin{equation}
\label{ns}
\mathbf{z} = \sqrt{\alpha(\mathbf{m})}\odot \mathbf{x} + \sqrt{1-\alpha(\mathbf{m})}\odot \epsilon, \ \ \epsilon \sim \mathcal{N}(0, \mathbf{I}),
\end{equation}
where $\mathbf{z}$ is the noised states and $\alpha: [0, K] \to [0, 1]$ maps noise level values to corresponding noise scales, where $K$ is the maximum noise level. During training, MDG samples noise masks across agent-time positions, and the denoiser $\mathcal{D}$ learns to reconstruct the clean trajectories from the corresponding noised inputs:
\begin{equation}
\small
\mathcal{L} = \mathbb{E}_{\mathbf{x},\mathbf{m},\epsilon} \left[ \|\mathcal{D}(\sqrt{\alpha(\mathbf{m})} \odot \mathbf{x} + \sqrt{1-\alpha(\mathbf{m})} \odot \epsilon, \mathbf{m}) - \mathbf{x}\|^2 \right],
\end{equation}
where $\mathbf{m} \sim p_{\text{mask}}$ is sampled from a predefined distribution. 

\noindent \textbf{Generation via Masked Denoising.} In inference, MDG follows a predefined denoising schedule $\{\bar{\mathbf{m}}_{L}, \bar{\mathbf{m}}_{L-1}, \ldots, \bar{\mathbf{m}}_{0}\}$, where $L$ is the number of denoising steps. Each mask $\bar{\mathbf{m}}_{\ell}$ defines the desired noise level for every spatiotemporal position (agent and timestep) at step $\ell$. The process begins from a fully noised state $\mathbf{z}_L \sim \mathcal{N}(0, \mathbf{I})$ corresponding to the highest noise mask $\bar{\mathbf{m}}_L$, and progressively reduces the noise level toward $\bar{\mathbf{m}}_0$, representing the clean output.

At step $\ell$, the model receives the current noisy state $\mathbf{z}_{\ell}$ and the mask $\bar{\mathbf{m}}_{\ell}$, and produces a clean estimate $\hat{\mathbf{x}}_{\ell} = \mathcal{D}(\mathbf{z}_{\ell}, \bar{\mathbf{m}}_{\ell}).$ If iterative refinement is desired, the clean estimate is then re-noised according to the next mask in the schedule for the subsequent denoising step:
\begin{equation}
\label{eq:renoise}
\mathbf{z}_{\ell-1} =
\sqrt{\alpha(\bar{\mathbf{m}}_{\ell-1})} \odot \hat{\mathbf{x}}_{\ell}
+ \sqrt{1 - \alpha(\bar{\mathbf{m}}_{\ell-1})} \odot \epsilon.
\end{equation}

At each iteration, the model predicts a clean reconstruction, which is then re-noised to the level specified by $\bar{\mathbf{m}}_{\ell-1}$ before the next call. After the final iteration, the model outputs $\mathbf{z}_{0}$, the fully denoised trajectory ready for use.

\begin{figure*}
    \centering
    \includegraphics[width=0.97\linewidth]{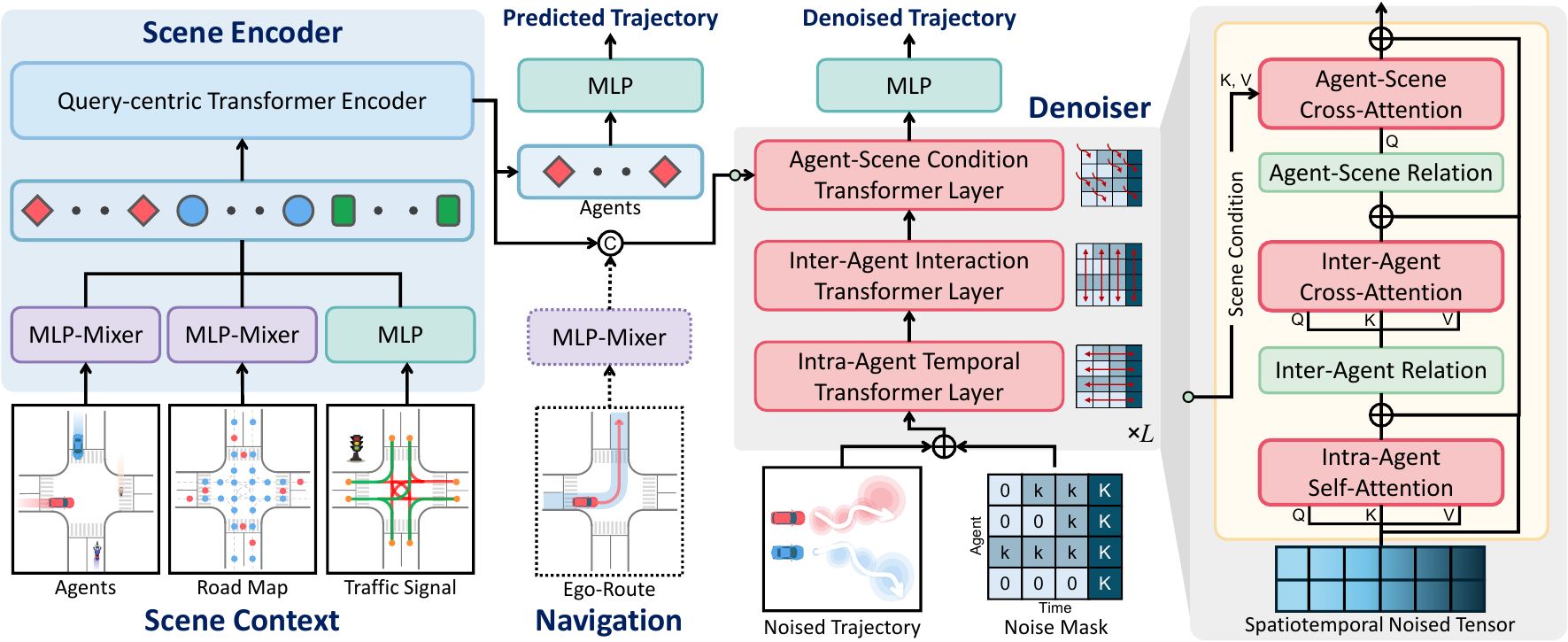}
    \caption{Overview of the MDG model structure. The Scene Encoder integrates scene context, including agent states, map polylines, and traffic lights, using modality-specific networks and a query-centric Transformer to produce a unified scene representation. An auxiliary MLP head decodes from the representation of agents to predict trajectories as regularization. The ego route polylines are encoded via an MLP-Mixer network. The Denoiser processes mask-conditioned noised spatiotemporal trajectory tensor through stacked Transformer blocks with: intra-agent temporal self-attention, inter-agent interaction cross-attention, and agent-scene condition cross-attention, where only the ego agent attends to its route context for planning tasks. A final MLP head outputs clean, denoised trajectories. }
    \label{fig:mdg}
    \vspace{-0.4cm}
\end{figure*}

\subsection{Problem Formulation}

We consider a multi-agent traffic scene involving $N$ agents observed over a past temporal window $H$. The historical trajectories of all agents are denoted by $\mathcal{A} = \{x_a^{t}\}_{a=1:N}^{t=-H:0}$, where $t=0$ corresponds to the current timestep. The environmental context, including high-definition map features and traffic signal states, is represented by $\mathcal{M}$ and $\mathcal{S}$, respectively. These components form the scene context $\mathbf{c} = \{\mathcal{A}, \mathcal{M}, \mathcal{S}\}$. Given $\mathbf{c}$, the goal is to predict the joint future trajectories of all agents over a horizon $T$: $\mathbf{x} = \{x_a^{t}\}_{a=1:N}^{t=1:T}$.
Formally, the model learns a conditional generative mapping $p_{\theta}(\mathbf{x} | \mathbf{c})$. This unifies multiple downstream tasks: \textbf{Open-loop prediction:} the model generates the entire future sequence $\mathbf{x}$ in a single or few denoising steps; \textbf{Multi-agent simulation:} only a short segment of the predicted multi-agent future is executed before re-querying the model with updated agent histories; \textbf{Closed-loop Planning:} the model conditions generation on constraint, allowing consistent ego-agent motion planning.

\subsection{Model Structure}

The MDG model is based on a Transformer encoder-decoder architecture, designed to denoise spatiotemporal tensors and fuse scene context for multi-agent trajectory generation. \cref{fig:mdg} provides an overview of the model structure, and a detailed description is provided as follows. Additional details are provided in the supplementary material.

\noindent \textbf{Scene Encoder}.
MDG encodes a multimodal scene context comprising agent states $\mathcal{A} \in \mathbb{R}^{N \times H \times D_a}$, vectorized map polylines $\mathcal{M} \in \mathbb{R}^{N_m \times N_w \times D_w}$, and traffic light states $\mathcal{S} \in \mathbb{R}^{N_s \times D_s}$.
For planning tasks, we additionally condition on the ego agent’s navigation route $\mathcal{R} \in \mathbb{R}^{N_r \times N_w \times D_w}$. Here, $N_m$, $N_r$, $N_w$, and $N_s$ denote the numbers of road map polylines, ego-route polylines, waypoints, and traffic lights, respectively; $D_a$, $D_w$, and $D_s$ represent their corresponding feature dimensions.
To process these inputs, MDG adopts three modality-specific MLP-Mixer encoders \cite{tolstikhin2021mlp} for agents, maps, and optional ego-routes, and an MLP encoder for traffic lights.
Temporal features in $\mathcal{A}$ and spatial features in $\mathcal{M}$ and $\mathcal{R}$ are aggregated through adaptive max-pooling along the temporal and waypoint axes, producing compact yet informative latent embeddings.
The resulting unified scene representation $\mathcal{C} \in \mathbb{R}^{(N + N_m + N_s) \times D}$ encodes all entities in a shared latent space.
Then, query-centric Transformer layers \cite{shi2024mtr++, huang2024versatile, zhou2023query} model high-order dependencies among agents, maps, and signals, yielding refined context features $\mathcal{C}' \in \mathbb{R}^{(N + N_m + N_s) \times D}$.
For the ego agent, route encoding is concatenated to preserve navigation intent.
To stabilize encoder learning, an auxiliary MLP head decodes $\mathcal{C}'$ to predict future trajectories, providing regularization signals.

\noindent \textbf{Denoiser}.
The denoiser operates on a noised future trajectory tensor $\mathbf{z} \in \mathbb{R}^{N \times T \times D_z}$ and its associated noise mask $\mathbf{m} \in \mathbb{N}_0^{N \times T}$, which specifies per-timestep, per-agent noise magnitudes.
Each token in $\mathbf{z}$ represents an action (acceleration, yaw rate), which is propagated through a differentiable motion model into continuous physical states $(x, y, \theta, v)$. The resulting state tensor is encoded with an MLP, while the noise mask is embedded separately and fused downstream.
The denoiser consists of stacked Transformer blocks that interleave three specialized attention mechanisms: (1) \textit{intra-agent temporal self-attention} to model temporal dependencies within each agent’s trajectory, (2) \textit{inter-agent interaction cross-attention} with learned inter-agent relation encoding to capture multi-agent interactions, and (3) \textit{agent-scene condition cross-attention} with agent-scene relation encoding to inject spatial and semantic context into trajectory refinement. All cross-attention modules employ relative relation encodings to preserve local coordinate invariance. For planning tasks, only the ego agent can access route-conditioned context. The stacked denoising layers progressively reconstruct clean trajectories, followed by an MLP decoder that outputs the final denoised trajectories.

\subsection{Training}

MDG is trained to reconstruct clean trajectories from arbitrarily noised inputs. The training objective jointly optimizes denoising and auxiliary prediction losses:
\begin{equation}
\mathcal{L} = \mathcal{L}_d + \lambda \mathcal{L}_p,
\end{equation}
where $\mathcal{L}_d$ is the denoising loss and $\mathcal{L}_p$ is the prediction loss, and $\lambda$ is the balance weight.

Given the noise mask $\mathbf{m}$ applied to the action sequence, the corrupted input is $\mathbf{z}$, and its corresponding physical states (derived through the differentiable dynamics model) are represented as $\mathbf{\hat{s}} = g(\mathcal{D}(\mathbf{z}, \mathbf{m}))$.
The per-sample denoising loss encourages the model to directly reconstruct clean states for all spatiotemporal positions:
\begin{equation}
\mathcal{L}_d = \frac{1}{N T} \sum_{i=1}^{N} \sum_{t=1}^{T} | \mathbf{\hat{s}}^{t}_{i} - \mathbf{s}^{t}_{i} |^2.
\end{equation}
Details of the auxiliary prediction loss $\mathcal{L}_p$ are provided in the supplementary material.

We employ a linear $\alpha$-scheduler that linearly distributes noise variance across discrete noise levels in the range $\alpha \in [0.99, 0.01]$. Higher noise levels correspond to larger variance (lower $\alpha$), while lower noise levels retain a stronger signal component. This scheduling ensures stable learning across denoising difficulty levels.

Training MDG with strong noise or random noise can obscure inter-agent dependencies and hinder learning. To address this, we introduce an adaptive masking strategy that varies the masking rate $\delta$ across samples in a batch and applies noise either along the temporal or agent dimension. For each training sample: if \textit{temporal masking}, a $\delta$ fraction of later timesteps for each agent is fully noised, while remaining timesteps receive progressively increasing random noise levels; if \textit{agent masking}, a $\delta$ fraction of agents is fully noised, and the rest are assigned lower, uniform noise levels across timesteps.
The masking rate $\delta$ is uniformly distributed across samples within a batch, ensuring diverse exposure to corruption patterns. This encourages the denoiser to generalize across both temporal degradation and inter-agent corruption, improving its ability to recover structured multi-agent dynamics during inference. Further details are provided in the supplementary material.

\subsection{Inference}
Our MDG model supports flexible inference through customized denoising strategies tailored to diverse downstream scenarios, as illustrated in \cref{fig:hero}.

\noindent\textbf{One-step Denoising}.
Given the highly conditional nature of traffic-agent interactions, MDG can generate realistic trajectories in a single denoising step ($L=1$). Starting from a fully noised mask, the model produces clean rollouts directly. This strategy is well-suited for closed-loop planning, offering better runtime efficiency.

\noindent\textbf{Denoising along Time}.
Denoising can proceed along the temporal axis, gradually reducing noise over time steps. The denoised granularity can be flexibly adjusted, with noise levels decreasing progressively for traversed timesteps and remaining consistent across agents. This approach is effective for open-loop forecasting, promoting diversity and iterative refinement of multi-agent futures.

\noindent\textbf{Denoising along Agent}.
Alternatively, denoising can be performed agent-wise, selectively reconstructing subsets of agents at each step. Noise levels are consistent across time for individual agents but vary between agents. This enables conditional behavior prediction and target-agent planning, facilitating interactive and controllable simulation.

\noindent\textbf{Long-horizon Guidance}.
MDG enables long-horizon control by perturbing modified trajectories (with certain objective functions) with small additive noise and re-denoising.
This approach efficiently refines trajectory adjustments while constraining future behaviors within desired bounds, outperforming guided diffusion methods \cite{zhong2023guided, jiang2023motiondiffuser}.

\noindent\textbf{Closed-loop Result Reuse}.
To ensure temporal consistency during continuous rollouts, MDG can reuse previous-step results by shifting the latest actions and adding small perturbations. This supports high-frequency planning and reduces distributional drift caused by accumulated errors.

\section{Experiments}

\begin{table*}[htp]
\centering
\small
\caption{Closed-loop Multi-Agent Simulation Results on the Waymo Sim Agents Benchmark. * denotes results from the 2024 benchmark.}
\vspace{-0.2cm}
\label{tab:wosac_results}
\setlength{\tabcolsep}{4mm}
\begin{tabular}{@{} 
      l| 
      c| 
      c| 
      c| 
      c| 
      c  
    @{}}
    \toprule
    \textbf{Agent Policy} &
    \begin{tabular}[c]{@{}c@{}}
        \textbf{Realism Meta}\\
        \textbf{Metric} ($\uparrow$)
    \end{tabular} &
    \begin{tabular}[c]{@{}c@{}}
        \textbf{Kinematic}\\
        \textbf{Metric} ($\uparrow$)
    \end{tabular} &
    \begin{tabular}[c]{@{}c@{}}
        \textbf{Interactive}\\
        \textbf{Metric} ($\uparrow$)
    \end{tabular} &
    \begin{tabular}[c]{@{}c@{}}
        \textbf{Map-based}\\
        \textbf{Metric} ($\uparrow$)
    \end{tabular} &
    \begin{tabular}[c]{@{}c@{}}
        \textbf{minADE}\\
        {[}m{]} ($\downarrow$)
    \end{tabular} \\
    \midrule
    VBD* \cite{huang2024versatile}          & 0.7200 & 0.4169          & 0.7819 & 0.7207 & 1.4743 \\
    BehaviorGPT* \cite{zhou2024behaviorgpt} & 0.7473 & 0.4333        & 0.7997 & 0.7636 & 1.4147 \\
    SMART-Large* \cite{wu2024smart}         & 0.7614  & 0.4786        & 0.8066 & 0.7682 & 1.3728 \\
    DRoPE* \cite{zhao2025drope}            & 0.7625 & 0.4779         & 0.8065 & 0.7715 & \textbf{1.2626} \\
    UniMM \cite{lin2025revisit}             & 0.7829  & 0.4914         & 0.8089 & 0.9161 & 1.2949 \\
    SMART-CLSFT \cite{zhang2024closed}      & 0.7846 & 0.4931 & 0.8106 & 0.9177 & 1.3065 \\
    TrajTok \cite{zhang2025trajtok}         & 0.7852 & 0.4887 & \textbf{0.8116} & \textbf{0.9207} & 1.3179 \\
    SMART-R1  \cite{pei2025advancing}       & \textbf{0.7858} & \textbf{0.4944} & 0.8110 & 0.9201 & 1.2885  \\ \midrule
    \textbf{MDG (1-step, closed-loop)}   & 0.7844 & 0.4928         & 0.8099 & 0.9183 & 1.3123 \\
    \bottomrule
\end{tabular}
\vspace{-0.4cm}
\end{table*}

\subsection{Experimental Setup}

\noindent \textbf{Datasets.}
For the simulation and prediction tasks, we employ the Waymo Open Motion Dataset (WOMD)~\cite{ettinger2021large}, which contains 486,995 training scenarios, each covering 9 seconds of agent trajectories with the corresponding map data. To enable closed-loop simulation, we use the Waymax Simulator~\cite{gulino2024waymax} for roll-out. For planning evaluation, we adopt the nuPlan dataset~\cite{karnchanachari2024towards}, which comprises approximately 1,300 hours of real-world driving data. During training, we include all scenario types from the nuPlan dataset while limiting each type to a maximum of 4,000 scenarios, resulting in a total of 176,218 training samples.

\noindent \textbf{Implementation Details.}
Our MDG model consists of six query-centric Transformer encoder layers and two denoiser blocks, each containing three Transformer layers. The hidden dimension is set to $D=256$, leading to a total of approximately 10 million parameters. We adopt a five-level noise schedule ($K=5$), where the schedule parameter $\alpha$ in \cref{ns} is linearly distributed from $0.99$ to $0.01$. Additional training details are provided in the supplementary material. 

\noindent \textbf{Overview.}
MDG adopts a \textit{unified formulation and modeling} for multi-agent behavior generation, enabling a single model to handle diverse tasks without task-specific adaptations. Its \textit{one-stage training} and \textit{versatile denoising} facilitate both efficient closed-loop rollout and diverse open-loop prediction. The following experiments demonstrate the \textit{generality and effectiveness} of MDG across a wide range of traffic behavior modeling tasks.

\subsection{Closed-loop Tasks}

\subsubsection{Multi-Agent Simulation}
\textbf{Task Description.}
The objective of this task is to generate 32 future trajectories for up to 128 agents per scenario, each spanning 8 seconds and conditioned on 1 second of historical context. Agent trajectories are produced in a closed-loop manner using our MDG model in one-step denoising mode, which directly predicts clean samples and executes them within the Waymax simulator. Simulations are performed with a replanning frequency of 1~Hz. We follow the official evaluation protocol of~\cite{montali2024waymo}, which includes metrics evaluating motion realism, agent interactions, map compliance, and displacement error (minADE). The overall realism meta-metric is then derived as the primary metric.

\begin{figure*}[ht]
    \centering
    \includegraphics[width=0.99\linewidth]{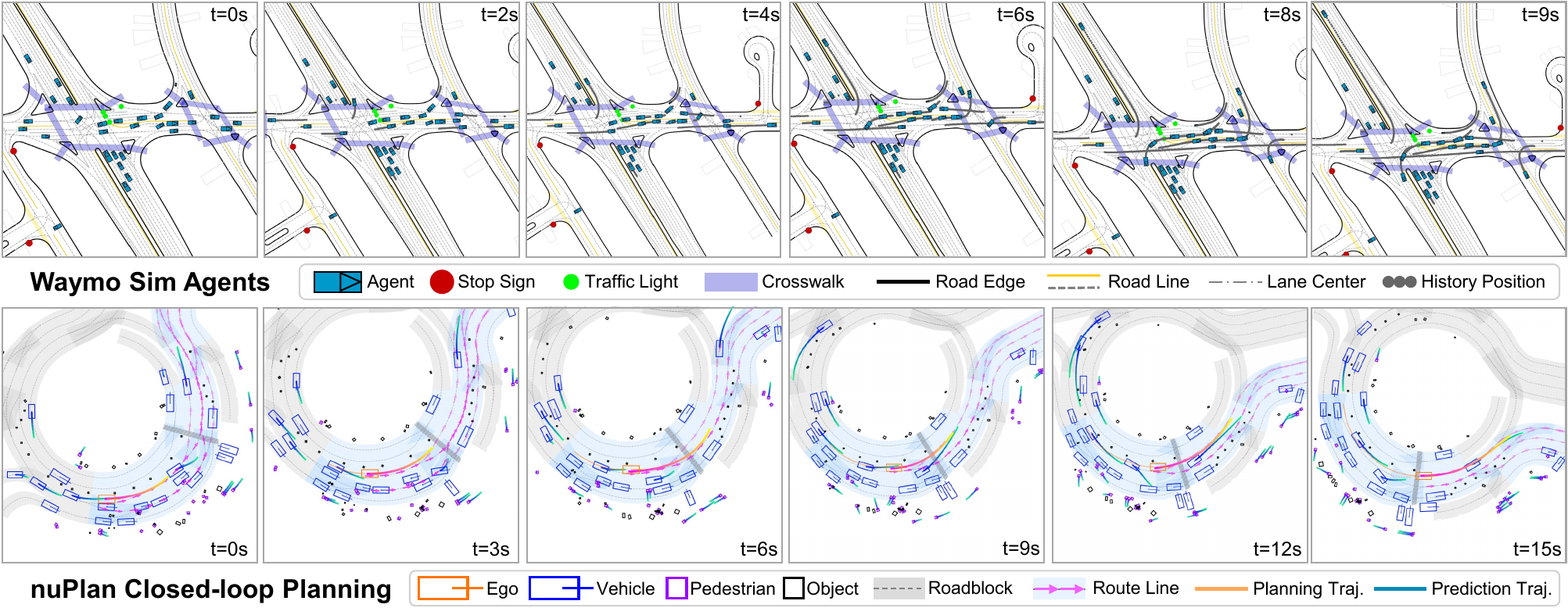}
    \caption{Qualitative results of MDG in closed-loop multi-agent simulation and ego-agent planning tasks. MDG controls all agents in an interactive, map-compliant manner and navigates the ego vehicle effectively in complex scenarios.}
    \label{fig:closedloop}
    \vspace{-0.4cm}
\end{figure*}

\noindent\textbf{Results.}
As shown in \cref{tab:wosac_results}, MDG achieves competitive performance on the Waymo Sim Agents Benchmark, closely matching the best results across core metrics, with only marginal differences ($<0.2\%$). Unlike SMART-based models (AR) such as SMART-R1 and SMART-CLSFT, which rely on multi-stage training and partially open-loop rollouts, MDG operates fully in closed-loop with a single-stage training. These results demonstrate that the proposed one-step denoising mode effectively captures multi-agent dynamics without requiring complex supervision or objectives. Additional closed-loop simulation results are provided in the supplementary material, and a qualitative example is shown in \cref{fig:closedloop}.

\subsubsection{Ego-Agent Planning}
\textbf{Task Description.}
This task evaluates the ability of navigating the ego agent along a predefined route in a closed-loop simulation environment. We evaluate on the nuPlan Val14~\cite{dauner2023parting} and Test14~\cite{cheng2024pluto} benchmarks under both non-reactive and reactive agent behavior modes. The \textit{closed-loop score (CLS)} is computed as the average across all scenarios, incorporating metrics such as Ego Progress, No At-Fault Collisions, and Drivable Area Compliance, where higher values indicate better performance. Each simulation scenario lasts 15 seconds and runs at 10~Hz.

\begin{table}[t]
\centering
\small
\caption{Closed-loop Motion Planning Results on the nuPlan Val14 and Test14 Benchmarks. NR and R denote simulation with non-reactive and reactive agent settings, respectively. * denotes methods that incorporate prior knowledge or rules about scoring. }
\vspace{-0.2cm}
\setlength{\tabcolsep}{2mm}
\begin{tabular}{l|cccc}
\toprule
\multirow{2}{*}{\textbf{Planner}} & 
\multicolumn{2}{c}{\textbf{Val14}} & 
\multicolumn{2}{c}{\textbf{Test14}} \\
\cmidrule(lr){2-3} \cmidrule(lr){4-5}
 & \textbf{NR} ($\uparrow$) & \textbf{R} ($\uparrow$) & \textbf{NR} ($\uparrow$) & \textbf{R} ($\uparrow$) \\
\midrule
IDM & 75.60 & 77.33 & 70.39 & 74.42 \\
GameFormer~\cite{huang2023gameformer} & 79.94 & 79.78 & 83.88 & 82.05 \\
PlanTF~\cite{cheng2024rethinking}  & 84.27 & 76.95 & 85.62 & 79.58 \\
PLUTO~\cite{cheng2024pluto} & 88.89 & 78.11 & 89.90 & 78.62 \\
Diffusion Planner~\cite{zheng2025diffusionbased} & 89.87 & 82.80 & 89.19 & 82.93 \\ 
PDM-Closed*~\cite{dauner2023parting} & \textbf{92.84} & \textbf{92.12} & 90.05 & \textbf{91.63} \\
CarPlanner*~\cite{Zhang_2025_CVPR}  & 91.45 & -- & \textbf{94.07} & 91.10 \\
\midrule
MDG (1-step) & 88.85 & 81.32 & 88.43 & 81.10 \\
\textbf{MDG (1-step, reuse)} & 90.45 & 83.89 & 90.16 & 83.21 \\
\bottomrule
\end{tabular}
\vspace{-0.4cm}
\label{nuplan}
\end{table}

\noindent\textbf{Results.}
As shown in \cref{nuplan}, MDG demonstrates strong closed-loop planning performance on the nuPlan benchmarks. The \textit{MDG (reuse 1-step)} variant achieves the best overall scores, surpassing the one-step version by selectively reusing ego-agent actions from the previous step, thereby enhancing trajectory consistency. Compared to previous diffusion-based planners, MDG attains higher scores, indicating improved performance in dynamic environments. While methods that incorporate explicit priors or rule-based scoring (\ie, CarPlanner and PDM-Closed) achieve marginally higher scores, MDG delivers comparable performance without relying on such heuristics, highlighting its generality and scalability.

\subsection{Open-loop Tasks}

\subsubsection{Multi-modal Motion Prediction}
\textbf{Task Description}.  
This task evaluates the model’s ability to predict six joint future trajectories for all agents in a scene, with a fixed prediction horizon of 8 seconds. Experiments are conducted on the WOMD validation split. We report three evaluation metrics: \textit{collision rate (CR)} to assess scene-level consistency of predictions; \textit{scene average displacement error (SADE)}, which quantifies the overall accuracy of the predicted trajectories; and \textit{minimum SADE (minSADE)}, which reflects the quality of the most accurate prediction. All metrics are computed over agents labeled as \textit{modeled}. The difference between SADE and minSADE can provide insight into the diversity of the prediction results. We also explore different inference modes for our MDG model, such as denoising along time and agent axes.

\begin{figure*}[ht]
    \centering
    \includegraphics[width=0.99\linewidth]{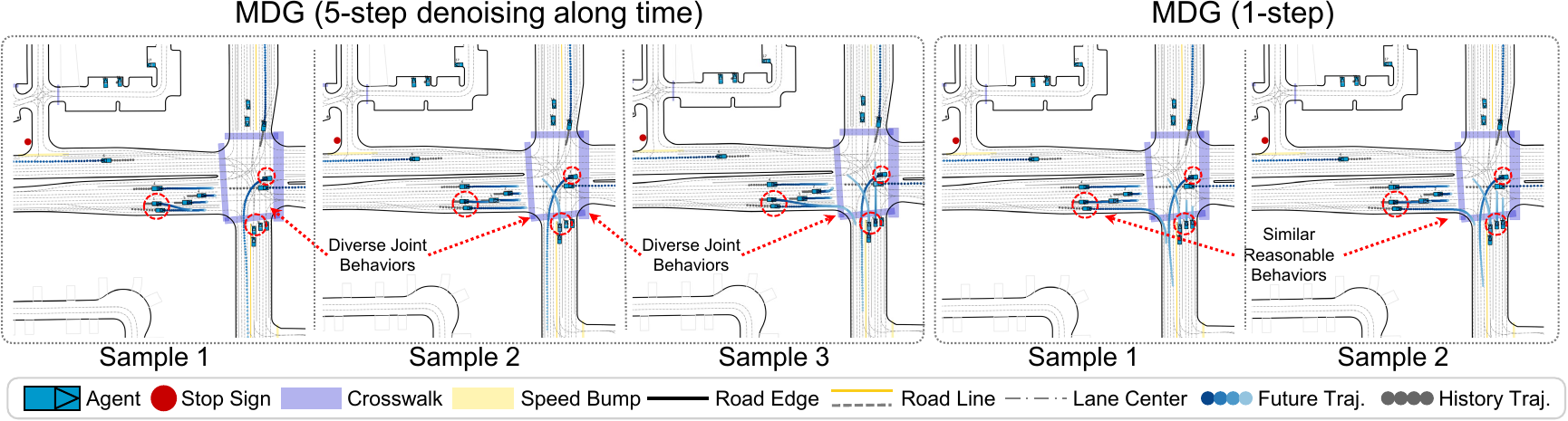}
    \caption{ Qualitative results of MDG in multi-agent open-loop prediction. The one-step denoising mode can produce plausible and interactive scenarios, but with limited sample diversity. By using multi-step denoising along the temporal axis, the generated scenarios exhibit greater diversity, and agents display obvious multimodal behaviors. }
    \label{fig:openloop}
    \vspace{-0.4cm}
\end{figure*}

\noindent\textbf{Results.}
Quantitative results in \cref{tab:pred} demonstrate that MDG achieves the best overall performance on the WOMD validation set. The MTR baseline shows the highest collision rate, indicating limited scene-level consistency due to marginal predictions. Both VBD and SMART achieve strong results, while MDG (1-step) achieves the lowest collision rate, and MDG (temporal, 5-step) achieves the lowest minSADE metric.
Increasing the number of denoising steps (along temporal or agent dimensions) generally enhances prediction diversity, as reflected by higher SADE and lower minSADE, although excessive denoising steps may still slightly degrade accuracy due to out-of-distribution sampling. Qualitative results in \cref{fig:openloop} further illustrate that MDG with five-step temporal denoising generates richer and more diverse multi-agent behaviors compared to the one-step setting. Additional results are provided in the supplementary material.

\begin{table}[t]
\centering
\small
\setlength{\tabcolsep}{1.3mm}
\caption{Open-loop Prediction Results on WOMD Validation Set}
\vspace{-0.2cm}
\begin{tabular}{l|ccc}
\toprule
\textbf{Method }                & \textbf{CR} [\%] ($\downarrow$)           & \textbf{SADE} ($\downarrow$)       & \textbf{minSADE} ($\downarrow$)  \\ \midrule
MTR  \cite{shi2024mtr++}        & 9.990             & 3.171         & 2.014 \\
VBD \cite{huang2024versatile}   & 6.427             & 2.988         & 2.006 \\ 
SMART \cite{zhang2024closed}    & 4.993             & 2.896         & 1.849 \\\midrule
MDG (1-step)                    & \textbf{4.415}    & \textbf{2.755}& 1.951  \\
MDG (time 5-step)               & 4.875             & 3.122         & \textbf{1.840}  \\
MDG (agent 5-step)              & 4.641             & 2.994         & 1.863\\
MDG (time 10-step)              & 5.026             & 3.233         & 1.964 \\
MDG (agent 10-step)             & 4.981             & 3.107         & 1.916 \\
MDG (time 20-step)              & 5.071             & 3.279         & 2.050 \\
\bottomrule
\end{tabular}
\label{tab:pred}
\vspace{-0.4cm}
\end{table}

\subsubsection{Controllable Generation}
\textbf{Task Description}.
This task evaluates controllable scenario generation, where the behaviors of target agents are conditioned on predefined goals. To evaluate both controllability and scene consistency, we adopt an open-loop generation setting and produce three samples with different random seeds. Specifically, ground-truth goals are assigned to the labeled target agents, and the model generates trajectories for all agents in the scene. For MDG with multi-step guidance, we apply denoising along the agent axis. We use the following evaluation metrics: \textit{collision rate (CR)}, \textit{SADE}, and \textit{goal-reach rate (GR)}. These metrics quantify how effectively the assigned agents achieve the desired target behaviors while preserving the coherence of the scene. We randomly select 1K scenarios from the WOMD validation set for evaluation. The detailed guidance procedure is provided in the supplementary material.

\begin{figure}
    \centering
    \includegraphics[width=0.95\linewidth]{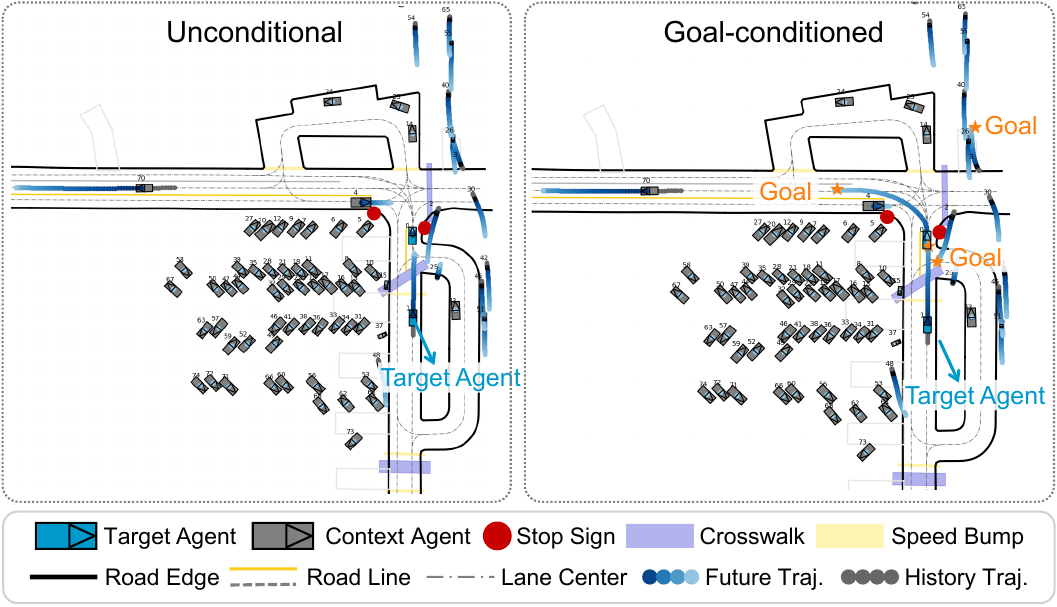}
    \caption{Illustration of the controllable generation task. Goals are assigned to target agents, and MDG guides them to reach the designated goals while maintaining reactions for surrounding agents.}
    \label{fig:result}
    \vspace{-0.4cm}
\end{figure}

\noindent \textbf{Results}.
The results in \cref{tab:control} indicate that the MDG model achieves better controllability and scene consistency compared to the guided VBD model. MDG attains higher goal-reach rates and lower collision rates, indicating a better balance between goal satisfaction and scene consistency. For the MTR model, directly selecting the trajectory closest to the target goal yields the highest GR and lowest SADE, but this heuristic approach significantly degrades scene consistency, as reflected in increased collision rates. Among different inference modes, the 5-step guidance offers the best trade-off between controllability and consistency, outperforming 1-step and 10-step settings. Runtime comparisons in \cref{tab:runtime} further show that MDG significantly reduces computation overhead compared to diffusion-based (VBD) guidance, highlighting its efficiency for controllable scenario generation.

\begin{table}[ht]
\centering
\small
\renewcommand{\arraystretch}{1.1}
\setlength{\tabcolsep}{1.1mm}
\caption{Results on Controllable Scenario Generation}
\vspace{-0.2cm}
\begin{tabular}{l|ccc}
\toprule
\textbf{Method }              & \textbf{CR} [\%] ($\downarrow$)                &\textbf{GR} [\%] ($\uparrow$)                   & \textbf{SADE} ($\downarrow$)  \\ \midrule
MTR (goal)           & 15.13$\pm$0.08          & \textbf{69.3}$\pm$0.05    & \textbf{2.386}$\pm$0.002 \\
VBD (guide)          & 9.83$\pm$0.09           & 43.8$\pm$0.06             & 3.406$\pm$0.006 \\ 
MDG (guide 1-step)   & \textbf{5.03}$\pm$0.20  & 36.4$\pm$0.16             & 2.886$\pm$0.002 \\
MDG (guide 5-step)   & 5.42$\pm$0.03           & 48.9$\pm$0.21             & 2.919$\pm$0.007 \\
MDG (guide 10-step)  & 6.39$\pm$0.26           & 48.5$\pm$0.23             & 3.357$\pm$0.001 \\
\bottomrule
\end{tabular}
\label{tab:control}
\vspace{-0.4cm}
\end{table}

\begin{table}[ht]
\centering
\small
\setlength{\tabcolsep}{1pt}
\renewcommand{\arraystretch}{1.1}
\caption{Comparison of Runtime Performance (ms, mean $\pm$ std)}
\vspace{-0.2cm}
\label{tab:runtime}
\begin{tabular}{l|cccc}
\toprule
\textbf{Model} & \textbf{AR} & \textbf{1-step} & \textbf{5-step} & \textbf{5-step Guidance} \\
\midrule
SMART & 253.8$\pm$26.9 & -- & -- & -- \\
VBD & -- & 282.2$\pm$41.8 & 1155.0$\pm$94.6 & 9076.6$\pm$62.3 \\
MDG & -- & 252.6$\pm$49.4 & 1115.2$\pm$45.1 & 1160.7$\pm$48.4 \\
\bottomrule
\end{tabular}
\vspace{-0.4cm}
\end{table}

\subsection{Ablation Study}
\textbf{Effect of Training Noise Masking.}
We investigate the effect of different noise-masking strategies during MDG training, including random, binary, and multi-level masking. All models are tested with 5-step denoising, and results are reported in \cref{tab:noise}.
When random noise levels are used, the model fails to capture consistent spatiotemporal patterns, leading to degraded performance.
Binary masking ($K{=}1$) shows suboptimal performance due to limited masking diversity, while excessive noise levels ($K{=}10$) add unnecessary complexity without performance gain. Our used configuration ($K{=}5$) achieves an effective balance between noise variability and testing performance. Additional ablation results are provided in the supplementary materials.

\begin{table}[ht]
\centering
\small
\setlength{\tabcolsep}{3pt}
\caption{Influence of Noise Masking on Open-loop Prediction}
\vspace{-0.2cm}
\label{tab:noise}
\begin{tabular}{l|ccc}
\toprule
\textbf{Noise Mask }       & \textbf{CR} [\%] ($\downarrow$)    & \textbf{SADE} ($\downarrow$)  & \textbf{minSADE} ($\downarrow$) \\
\midrule
Random            & 12.477                          & 4.085                 & 3.982               \\
Multi $K=10$      & 4.956                           & 3.274                 & 1.956         \\
Multi $K=5$       & \textbf{4.875}                  & \textbf{3.122}        & \textbf{1.840}         \\
Binary $K=1$      & 5.635                           & 3.374                 & 2.222 \\
\bottomrule
\end{tabular}
\vspace{-0.4cm}
\end{table}

\section{Conclusions}
We introduce MDG, a masked denoising generative model for multi-agent behavior modeling in traffic scenarios. MDG leverages noise-based masking in a multi-agent spatiotemporal tensor and a Transformer-based denoiser to reconstruct clean samples from arbitrarily masked inputs. MDG achieves competitive performance in closed-loop simulation on the Waymo Sim Agents Benchmark and in planning on the nuPlan benchmark. MDG also proves effective in open-loop tasks such as trajectory prediction and goal-conditioned controllable generation. Future work will focus on improving runtime efficiency and extending MDG to integrate user-provided text or other conditioning inputs. 

{
    \small
    \bibliographystyle{ieeenat_fullname}
    \bibliography{main}
}

\clearpage
\maketitlesupplementary
\setcounter{figure}{0}
\setcounter{table}{0}
\setcounter{equation}{0}
\setcounter{section}{0}

\renewcommand{\thesection}{\Alph{section}}
\renewcommand{\thefigure}{S\arabic{figure}}
\renewcommand{\thetable}{S\arabic{table}}
\renewcommand{\theequation}{S\arabic{equation}}

\section{Model Details}
\subsection{Scene Encoder}
\textbf{Agent Encoder.}
The agent state tensor contains per-timestep states for all agents, including their $(x, y)$ coordinates, heading, velocity, and bounding box dimensions (length, width, height). All agent trajectories are transformed into their respective local coordinate frames, using the final observed state as the origin.
An MLP-Mixer encoder is employed to capture both temporal dynamics and feature interactions within each agent’s state sequence. Specifically, the agent state tensor is processed by stacked MLP-Mixer blocks that alternately mix information along the temporal and feature dimensions, producing a latent tensor of shape $[N, H, D]$, where $N$ is the number of agents, $H$ the historical timesteps, and $D$ the hidden dimension. Temporal max-pooling is then applied to summarize motion patterns across time. Each agent type is embedded via a learnable embedding vector and added to its feature representation. The final agent encoding has the shape $[N, D]$.

\noindent \textbf{Map Encoder.}
The map tensor consists of $N_m$ polylines (e.g., road centerlines, lane boundaries, crosswalks), each with $N_w$ waypoints described by $(x, y)$ coordinates and heading. All map elements are transformed into local coordinates using the first waypoint of each polyline as the origin.
Each polyline is processed by an MLP-Mixer that models both intra-polyline spatial dependencies and cross-feature correlations. The Mixer operates on a tensor of shape $[N_m, N_w, D]$, and its outputs are aggregated by max-pooling along the waypoint axis to produce a polyline-level feature tensor $[N_m, D]$. Polyline type and associated traffic-signal state are encoded via learnable embeddings and added to the feature representation. The final map encoding has the shape $[N_m, D]$.

\noindent \textbf{Ego-Route Encoder.}
For planning tasks, we include route polylines for the ego agent, consisting of $N_r$ polylines, each with $N_w$ waypoints defined by $(x, y)$ coordinates and heading. All waypoints are transformed into local coordinates using the first waypoint of each polyline as the origin.
An MLP-Mixer encoder processes these route polylines in the same manner as the map encoder, capturing both waypoint-wise spatial relations and inter-feature dependencies. The resulting ego-route encoding has shape $[N_r, D]$.

\noindent \textbf{Traffic Signal Encoder.}
Traffic signals are represented by their stop points. Since we adopt relative spatial encoding, only the current signal phase (e.g., red, yellow, green) is encoded. An MLP Embedding layer converts each signal state into a feature tensor of shape $[N_s, D]$.

\noindent \textbf{Relative Relation Encoder.}
To model spatial relationships among all scene elements, we compute pairwise relative attributes between every pair $(i, j)$. The relative distance and heading are computed as follows: (1) for agents, using the last observed position; (2) for map elements, using the first waypoint of each polyline; and (3) for traffic signals, using the stop point position. Self-relations are assigned a small constant.
An MLP-based Fourier Embedding relation encoder~\cite{zhou2023query} encodes these pairwise relations into a relation tensor of shape $[N+N_m+N_s, N+N_m+N_s, D]$, enabling subsequent attention modules to incorporate spatial context and topology-aware reasoning.

\noindent \textbf{Query-centric Transformer Encoder.}
The Transformer encoder is employed to extract relationships among all scene elements. It comprises a stack of query-centric attention layers, which use the same attention mechanism in \cref{qc}. The relative relation encodings are incorporated into this encoder. All scene elements (agents, map polylines, and traffic signals) are concatenated into a tensor of shape $[N+N_m+N_s, D]$, which is input to the Transformer. After passing through several attention layers, the Transformer learns to capture the interactions and relationships between the scene elements. Invalid elements are masked out during the attention calculations. The final scene context encoding retains its original shape $[N+N_m+N_s, D]$. The relative position and heading differences between elements $i$ and $j$ are encoded as edge attributes, forming a relation encoding tensor $\mathbf{e}^{i \rightarrow j}$. The relative cross-attention (RCA) is defined as:
{
\small
\begin{align}
\label{qc}
RCA(Q^i, K, V, \mathbf{e}) 
&= \text{softmax} \Bigg( 
\frac{\mathbf{q}^i}{\sqrt{D}} 
\Bigg[ 
\Big\{ \mathbf{k}^j + \mathbf{e}^{i \rightarrow j} \Big\}_{j \in \Omega(j)} 
\Bigg]^T 
\Bigg) \nonumber \\
&\quad \times 
\Bigg( 
\Big\{ \mathbf{v}^j + \mathbf{e}^{i \rightarrow j} \Big\}_{j \in \Omega(j)} 
\Bigg),
\end{align}
} 

\noindent where $\mathbf{q}^i, \mathbf{k}^j, \mathbf{v}^j$ represent the query, key, and value elements respectively, each containing relevant element-centric information, and $j \in \Omega(j)$ defines the set of indices corresponding to neighboring elements.

\subsection{Decoder}
\noindent \textbf{Noised Future Encoder.}
We first compute each agent’s control actions (acceleration and yaw rate) from their logged trajectories. These actions are normalized to form a clean action tensor of shape $[N, T_a, 2]$, where $T_a$ denotes the reduced temporal dimension obtained via action chunking, which improves computational and memory efficiency.
Gaussian noise is then applied to the clean actions, which are subsequently transformed into noisy physical states (positions, headings, and velocities) through a differentiable dynamics function, resulting in a tensor of shape $[N, T_a, 5]$. The noised states are encoded using an MLP. Mask and timestep embeddings are then added to produce the noised future query tensor of shape $[N, T_a, D]$, serving as the input to the denoising Transformer.

\noindent \textbf{Transformer Denoiser.}
The denoiser consists of stacked Transformer blocks, each containing complementary attention layers that operate over different relational dimensions:
\begin{itemize}
\item \textbf{Intra-Agent Temporal Attention.} A multi-head self-attention layer extracts temporal dependencies within each agent’s future trajectory.
\item \textbf{Inter-Agent Interaction Attention.} The outputs are passed through a cross-attention layer that models inter-agent interactions. Invalid or missing agents are masked. Relation encodings from the encoder are used as positional and relational priors (inter-agent relations) in this attention computation.
\item \textbf{Agent-Scene Condition Cross-Attention.} A cross-attention layer fuses the agent features with scene context representations from the encoder (map, traffic lights, and agents). For planning tasks, the ego agent receives additional conditioning through cross-attention with the ego-route encoding, which provides route-level geometric and semantic guidance. This route information is exclusively accessible to the ego agent. Relation encodings for agents and scene elements (map polylines, traffic agents, traffic lights, and optional ego-route polylines) are incorporated into this attention process.
\end{itemize}
Residual connections are applied across all Transformer layers to stabilize optimization and preserve gradient flow. The final denoised latent tensor is decoded with an MLP that outputs denoised action sequences for each agent.

\noindent \textbf{Trajectory Decoder.}
An auxiliary MLP decoder operates directly on the agent-specific scene context encodings to predict future trajectories of shape $[N, M, T, 3]$, where $M$ is the number of trajectory modalities. This auxiliary prediction branch stabilizes training and encourages the scene encoder to learn trajectory-relevant representations.

\noindent \textbf{Differentiable Dynamic Function.}
We adopt the differentiable dynamics function from~\cite{huang2024versatile} to convert predicted agent actions (acceleration and yaw rate) into corresponding physical states (positions and headings), conditioned on each agent’s current state. As this function is differentiable, it is integrated into the model as a trainable layer, enabling end-to-end gradient propagation through both the kinematic transformation and the denoising process.

\subsection{Model Parameters}
The primary parameters used in our MDG model are summarized in \cref{modelpara} and \cref{modelpara-nuplan}.
Separate configurations are adopted for the Waymo and nuPlan experiments to account for differences in scene complexity and task requirements. Unless otherwise specified, parameters in the nuPlan setup are identical to those in the Waymo configuration.

\begin{table*}[ht]
\centering
\caption{Model Parameters for Waymo Experiments}
\small
\begin{tabular}{lcc}
\toprule
\textbf{Parameter} & \textbf{Value} & \textbf{Description} \\
\midrule
$N$ & 128 & Number of agents per scene \\
$N_m$ & 320 & Number of map polylines \\
$N_w$ & 16 & Number of waypoints per polyline \\
$N_s$ & 16 & Number of traffic lights \\
$H$ & 11 & History steps \\
$T$ & 80 & Future steps \\
$T_a$ & 40 & Reduced future timesteps \\
$D$ & 256 & Hidden dimension \\
$M$ & 6 & Number of trajectory modalities in predictor \\
\midrule
Modality encoder layers & 2 & Number of MLP-Mixer layers in the encoder \\
Encoder layers & 6 & Number of query-centric Transformer layers \\
Decoder blocks & 2 & Number of Transformer denoising blocks \\
Attention heads & 8 & Number of attention heads per layer \\
Dropout rate & 0.1 & Dropout probability \\
Feed-forward dimension & 1024 & Dimension of the feed-forward network \\
Action chunk & 2 & Action sequence granularity \\
Action mean & $[0.0, 0.0]$ & Mean of action distribution [acceleration, yaw rate] \\
Action std & $[1.0, 0.5]$ & Standard deviation of action distribution [acceleration, yaw rate] \\
\bottomrule
\end{tabular}
\label{modelpara}
\end{table*}

\begin{table*}[ht]
\centering
\caption{Model Parameters for nuPlan Experiments}
\small
\begin{tabular}{lcc}
\toprule
\textbf{Parameter} & \textbf{Value} & \textbf{Description} \\
\midrule
$N$ & 32 & Number of agents per scene \\
$N_m$ & 256 & Number of map polylines \\
$N_m$ & 32 & Number of ego-route polylines \\
$N_w$ & 20 & Number of waypoints per polyline \\
$H$ & 20 & History steps \\
$T$ & 80 & Future steps \\
\bottomrule
\end{tabular}
\label{modelpara-nuplan}
\end{table*}

\section{Training Details}
 The training objective combines two loss components and invalid agents and timesteps are excluded:
\begin{equation}
\mathcal{L} = \mathcal{L}_d + \lambda \mathcal{L}_p,
\end{equation}
where $\mathcal{L}_d$ represents the denoising loss, and $\mathcal{L}_p$ is an auxiliary prediction loss, and $\lambda=5$ is the balance weight. The prediction loss $\mathcal{L}_p$ is defined as:
\begin{equation}
\mathcal{L}_p = \frac{1}{N} \sum_{i=1}^{N} \mathcal{SL}_1( \mathbf{\hat s}^{*}_{i}  - \mathbf{s}^{gt}_{i}), \, {i}^{*} = \arg \min_{m} \| \mathbf{\hat s}^{m}_{i} - \mathbf{s}^{gt}_{i} \|_2,
\end{equation}
where $\mathcal{SL}_1$ denotes the smooth $L_1$ loss applied to the best-predicted trajectory $\mathbf{\hat s}^{*}_{i}$, defined as the trajectory with the minimal $L_2$  from the ground truth $\mathbf{s}^{gt}_{i}$.

\begin{figure}[ht]
    \centering
    \includegraphics[width=\linewidth]{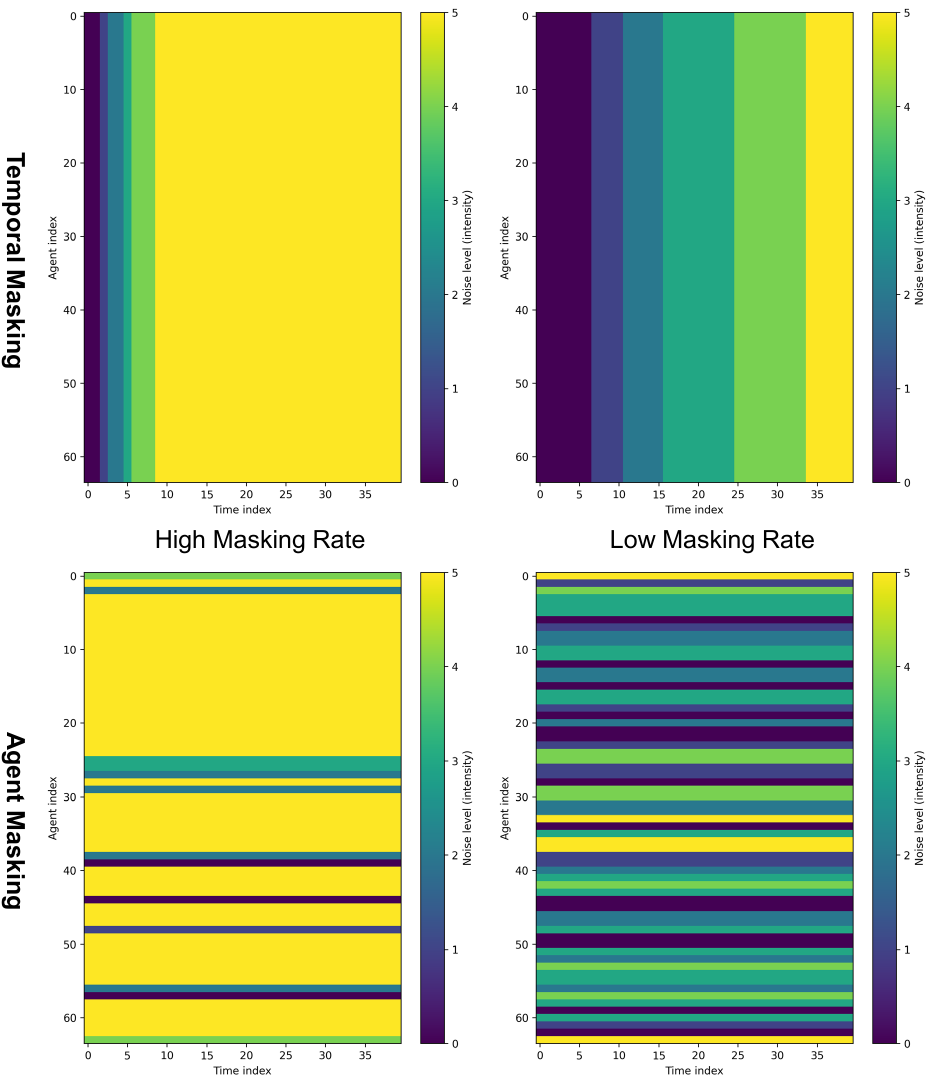}
    \caption{Illustration of the proposed training noise masking strategy. Top: Temporal masking; Bottom: Agent masking. Each column shows an example under either a high or low masking rate.}
    \label{fig:train_mask}
    \vspace{-0.3cm}
\end{figure}

We propose a novel masking strategy that distributes the masking rate across training batches and applies masking randomly along either the temporal or agent axis. This approach facilitates adaptive denoising generation during inference and guidance. The training procedure is detailed in \cref{alg:train}. The strategy involves randomly selecting masking patterns: masking over time or masking over agents. For each training sample, a masking rate $\delta$ is assigned. When masking over time, a $\delta$ fraction of the later timesteps for each agent are fully noised, while other timesteps are randomly assigned noise levels that progressively increase. When masking over agents, a $\delta$ fraction of the agents are fully noised with the highest noise levels, and the remaining agents are randomly assigned lower noise levels, while the noise levels for an individual agent remain consistent across timesteps. Importantly, the masking rate $\delta$ is evenly distributed across all samples in a batch to ensure robust learning. \cref{fig:train_mask} illustrates both masking types under high and low masking rates.

\begin{algorithm}
\caption{Training Procedure for MDG}
\centering
\label{alg:train}
\begin{algorithmic}[1]
\STATE {\bfseries Input:} Data $X$, Noise schedule $\alpha$, Noise level $K$, Denoising model $\mathcal{D}_\theta$,
\STATE {\bfseries Output:}  Trained model $\mathcal{D}_{\theta^*}$

\STATE Initialize model parameters $\theta$
\FOR{Each training batch $X_b$}
    \FOR{Each sample $x \in X_b$}
        \STATE Randomly select masking type: time-axis or agent-axis
        \STATE Assign masking rate $\delta \in [0,1]$
        \IF{masking over time}
            \STATE Apply full noise to $\delta$ fraction of later timesteps
            \STATE Assign random progressively increasing noise levels to remaining timesteps
        \ELSE
            \STATE Apply full noise to $\delta$ fraction of agents
            \STATE Randomly assign noise levels across timesteps for each agent
        \ENDIF
        \STATE Generate noise mask $\mathbf{m}$ and add noise to $x$ according to schedule $\alpha(\mathbf{m})$
    \ENDFOR
    \STATE Perform a forward pass of the model on $X_b$
    \STATE Compute loss $\mathcal{L}$
    \STATE Backpropagate the loss and update $\theta$ using an optimizer
\ENDFOR
\end{algorithmic}
\end{algorithm}

For Waymo experiments, to balance computational efficiency and model performance, we limit training to the 64 agents nearest to the labeled self-driving car (including itself). This reduces GPU memory usage while accommodating most scenarios, as they typically involve fewer than 64 valid agents. During testing, the model can scale up to 128 agents, leveraging the flexible attention mechanisms of the Transformer architecture. 

For the nuPlan experiments, following established practice \cite{cheng2024pluto, zheng2025diffusionbased}, we apply random perturbations to the current state as data augmentation. A short interpolation segment is inserted to ensure a physically plausible transition, allowing the model to remain robust to perturbations and converge back to the ground-truth trajectory.

Training is performed on eight NVIDIA L40S GPUs using the AdamW optimizer with a weight decay of 0.01 and BFloat16 precision. The set of training hyperparameters is provided in \cref{tab:hyperparameters}.

\begin{table*}[ht]
  \centering
  \small
  \caption{Hyperparameters for Model Training}
  \label{tab:hyperparameters}
  \begin{tabular}{lcc}
    \toprule
    \textbf{Hyperparameter} & \textbf{Value} & \textbf{Description} \\
    \midrule
    Noise levels & 5 & Number of distinct noise levels \\
    Learning rate & 0.0002 & Initial learning rate \\
    LR decay step  & 2000 & Step interval for learning rate decay \\
    LR warmup step & 1000 & Warmup steps at the beginning \\
    LR decay factor & 0.98 & Multiplicative decay factor \\
    Batch size & 4 & Number of samples per GPU \\
    Training epochs & 20 & -- \\
    Gradient clip & 1.0 & Maximum gradient norm for clipping \\
    \bottomrule
  \end{tabular}
\end{table*}

\section{Inference Details}
\textbf{Multi-step Denoising.}
Our proposed MDG framework offers flexibility during inference by enabling multi-step generation along either the temporal axis or the agent axis. Below, we elaborate on the inference process of our masked denoising model. The inference begins by initializing the process with full Gaussian noise, corresponding to a mask $\mathbf{\bar{m}}_L$ that fully covers the data with noise. This noisy input, along with the corresponding mask, is fed into the denoising model to generate an intermediate clean sample. In the subsequent iteration, the generated clean sample is re-noised according to the noise mask at the next step $\mathbf{\bar{m}}_{L-1}$. This noised sample is then passed through the model for denoising. This iterative process continues until the final step, where the final denoised sample is obtained. The complete inference procedure is illustrated in \cref{alg:inference}.

\begin{algorithm}
\caption{Inference Procedure for MDG}
\centering
\label{alg:inference}
\begin{algorithmic}[1]
\STATE {\bfseries Input:} Denoising step $L$, Noise mask schedule $\{\mathbf{\bar{m}}_\ell\}_{\ell=L}^0$, Noise schedule $\alpha$, Denoising model $\mathcal{D}_\theta$
\STATE {\bfseries Output:} Denoised sample $\mathbf{\hat x}=\mathbf{z}_{0}$

\STATE Initialize $\mathbf{z}_L \sim \mathcal{N}(0, \mathbf{I})$ \COMMENT{Start with full Gaussian noise}
\FOR{$\ell = L$ to $1$}
    \STATE $\mathbf{\hat x}_\ell \gets \mathcal{D}_\theta(\mathbf{z}_\ell, \mathbf{\bar{m}}_\ell)$ \COMMENT{Denoise the current sample}
    \STATE $\mathbf{z}_{\ell-1} \gets \sqrt{\alpha(\mathbf{\bar{m}}_{\ell-1})} \mathbf{\hat x}_\ell + \sqrt{1-\alpha(\mathbf{\bar{m}}_{\ell-1})} \epsilon, \ \epsilon \sim \mathcal{N}(0, \mathbf{I})$ \COMMENT{Re-noise for next step}
\ENDFOR

\STATE {\bfseries Return} $\mathbf{z}_{0}$ \COMMENT{Final denoised output}

\end{algorithmic}
\end{algorithm}

\begin{figure}[ht]
    \centering
    \includegraphics[width=\linewidth]{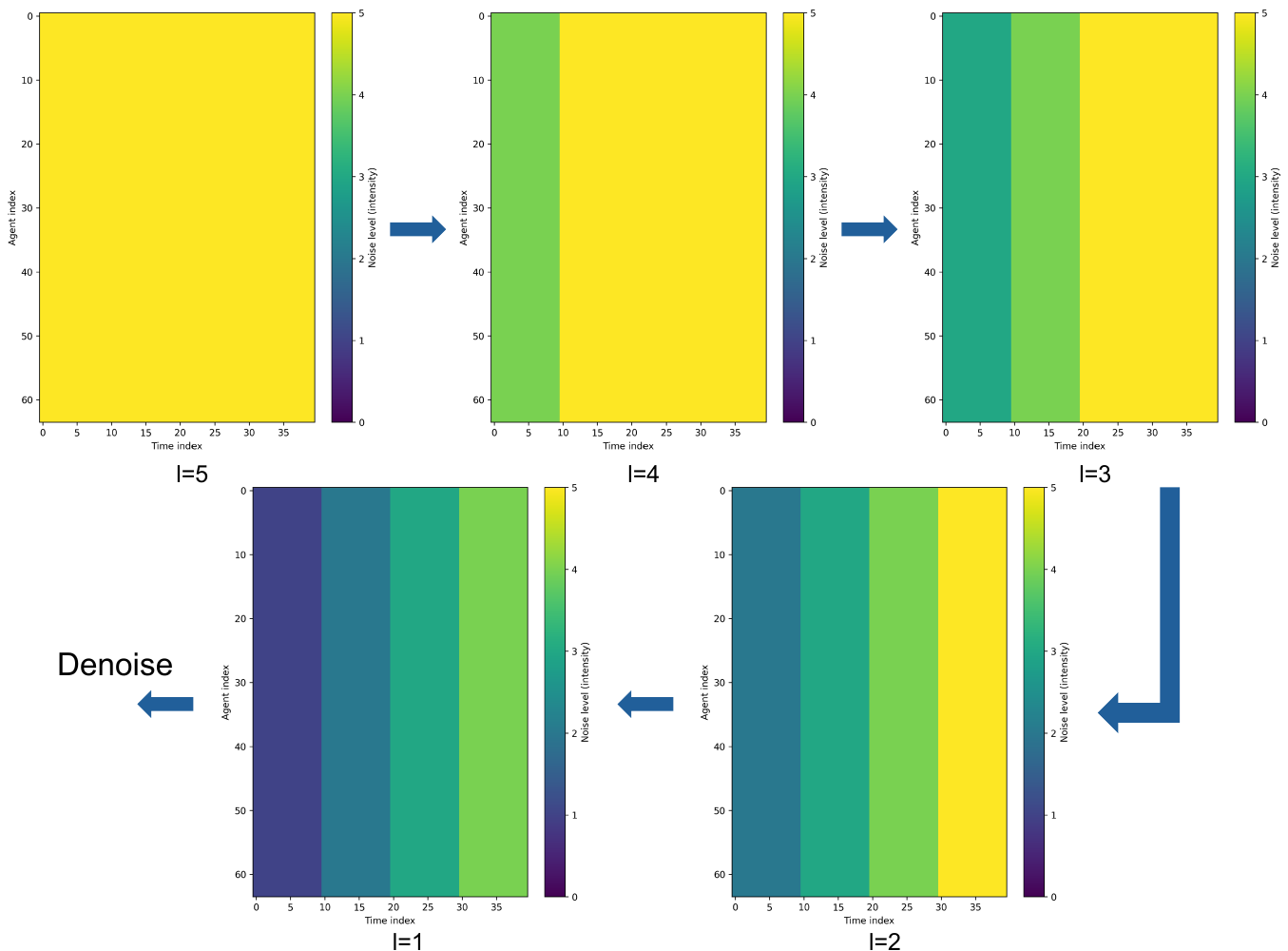}
    \caption{Example of a five-step temporal-axis denoising schedule used during inference.}
    \label{fig:infer_mask}
\end{figure}

The design of denoising schedules at inference time is important. We propose three strategies for this purpose. 1) \textit{Single-step Denoising:} The model produces a clean output in a single pass starting from a fully masked noisy input. 2) \textit{Temporal Axis Denoising:} Denoising proceeds progressively along the temporal dimension, analogous to autoregressive next-step generation. The noise level at each step is adjusted based on the remaining timesteps and the predefined noise scale, enabling flexible and fine-grained temporal refinement. 3) \textit{Agent Axis Denoising:} Similar to temporal-axis denoising, this strategy iterates along the agent dimension instead. It offers flexibility in the ordering and progression of agents, which can be advantageous in multi-agent scenarios with heterogeneous importance or interaction structures. \cref{fig:infer_mask} illustrates an example of the temporal-axis inference schedule. These flexible strategies allow the MDG framework to tailor its denoising process to diverse tasks and operational constraints.


\noindent \textbf{Guidance Imposition.}
The training mechanism of the MDG model facilitates a straightforward yet effective guidance imposition method to change the behavior of selected target agents while preserving the reactivity of other agents and maintaining overall scene consistency. This method involves replacing the denoised trajectories of the target agents with modified ones and adding small controlled noise. The adjusted trajectories are then fed into the next denoising iteration. To implement this, we define a guidance noise mask $\mathbf{\bar{g}}$, assigning a fixed low noise level $\alpha=0.8$ to all timesteps of the target agents while leaving other agents unaffected. This mask remains consistent throughout the denoising generation process, and the noise mask schedule controls the denoising of other agents. This design allows us to explicitly optimize the trajectories of the target agents or replace them with prior prediction results, without compromising the overall scene dynamics. The guidance procedure is illustrated in \cref{alg:guidance}. Notably, unlike diffusion-based models \cite{jiang2023motiondiffuser, zhong2023guided}, our method avoids differentiating through the denoiser, which shows significantly faster inference speeds, enhancing its practical applicability.

\begin{algorithm}
\caption{Guidance Imposition Procedure for MDG}
\centering
\label{alg:guidance}
\begin{algorithmic}[1]
\STATE {\bfseries Input:} Noise mask schedule $\{\mathbf{\bar{m}}_\ell\}_{\ell=L}^0$, Guidance noise mask $\mathbf{\bar g}$, Denoising schedule $\alpha$, Denoising model $\mathcal{D}_\theta$, Objective function $\mathcal{J}$
\STATE {\bfseries Output:} Denoised sample $\mathbf{\hat x}=\mathbf{z}_{0}$

\STATE Initialize $\mathbf{z}_L \sim \mathcal{N}(0, \mathbf{I})$ \COMMENT{Start with full Gaussian noise}
\FOR{$\ell = L$ to $1$}
    \STATE $\mathbf{\hat x}_\ell \gets \mathcal{D}_\theta(\mathbf{z}_\ell, \mathbf{\bar{m}}_\ell)$ \COMMENT{Denoise the current sample}
    \STATE $\mathbf{\hat x}_\ell^M \gets \mathcal{J}(\mathbf{\hat x}_\ell)$ \COMMENT{Modify the target agents' trajectories on the clean sample}
    \STATE $\mathbf{z}_{\ell-1} \gets \sqrt{ \alpha(\mathbf{\max(\bar{m}}_{\ell-1}, \mathbf{\bar g}))}  \mathbf{\hat x}_\ell^M + \sqrt{1-\alpha(\mathbf{\max(\bar{m}}_{\ell-1}, \mathbf{\bar g}))}  \epsilon, \ \epsilon \sim \mathcal{N}(0, \mathbf{I})$ \COMMENT{Apply guidance mask and re-noise for the next step}

\ENDFOR

\STATE {\bfseries Return} $\mathbf{z}_0$ \COMMENT{Final denoised output}

\end{algorithmic}
\end{algorithm}

\begin{table*}[ht]
\centering
\small
\caption{Influence of the auxiliary trajectory predictor on closed-loop performance}
\setlength{\tabcolsep}{5pt}
\begin{tabular}{lccc}
\toprule
\textbf{Task} & \textbf{Predictor} & \textbf{Main Metric ($\uparrow$)} & \textbf{minADE (m) ($\downarrow$)} \\
\midrule
\multirow{2}{*}{Waymo Sim Agents} 
& \checkmark & \textbf{0.7842} (Realism Meta) & \textbf{1.3123} \\
& --          & 0.7682                       & 1.4117 \\
\midrule
\multirow{2}{*}{nuPlan Val14-NR} 
& \checkmark & \textbf{89.75} (Closed-loop Score) & -- \\
& --          & 86.90                & -- \\
\bottomrule
\end{tabular}
\label{tab:predictor_influence}
\end{table*}

\begin{table*}[ht]
\centering
\small
\caption{Influence of action-to-state conversion on open-loop multi-agent trajectory prediction}
\begin{tabular}{l|cccc}
\toprule
Method              &  CR (\%) $\downarrow$       &  OR (\%) $\downarrow$   & SADE (m) $\downarrow$      & minSADE (m) $\downarrow$  \\ \midrule
W/ conversion       & \textbf{4.875}  & \textbf{3.274}   & \textbf{3.122}& \textbf{1.840}  \\
W/o conversion      & 8.875           & 6.066            & 4.353         & 2.853 \\
\bottomrule
\end{tabular}
\label{tab:physical}
\end{table*}

\begin{table*}[ht]
\centering
\small
\caption{Influence of model scale on open-loop multi-agent trajectory prediction}
\begin{tabular}{lc|cccc}
\toprule
Scale            & \# Model Params        &  CR (\%) $\downarrow$     &  OR (\%) $\downarrow$   & SADE (m) $\downarrow$      & minSADE (m) $\downarrow$ \\ \midrule
D=128               &  2.5M                  & 5.477            & 4.585             & 3.714            &  2.358 \\
D=256               &  10.0M                 & \textbf{4.875}   & \textbf{3.274}    & \textbf{3.122}    & \textbf{1.840}  \\
D=512               &  39.4M                 & 4.905            & 3.425             & 3.289             & 1.921\\ 
\bottomrule
\end{tabular}
\label{tab:scale}
\end{table*}

\begin{table*}[ht]
\centering
\caption{Ablation on attention mechanisms in the MDG denoiser on open-loop multi-agent trajectory prediction}
\small
\setlength{\tabcolsep}{4.5pt}
\begin{tabular}{ccc|ccc}
\toprule
\multicolumn{3}{c|}{\textbf{Attention Modules}} & \multicolumn{3}{c}{\textbf{Metrics}} \\
\cmidrule(lr){1-3} \cmidrule(lr){4-6}
Intra-Agent & Inter-Agent & Agent-Scene & CR (\%) $\downarrow$ & SADE (m) $\downarrow$ & minSADE (m) $\downarrow$ \\
\midrule
\checkmark & \checkmark & \checkmark & \textbf{4.875} & \textbf{3.122} & \textbf{1.840} \\
\checkmark & --          & \checkmark & 6.821 & 3.580 & 2.060 \\
--          & \checkmark & \checkmark & 5.612 & 3.460 & 1.990 \\
--          & --          & \checkmark & 9.964 & 4.070 & 2.310 \\
\bottomrule
\end{tabular}
\label{tab:attn_ablation}
\end{table*}

\begin{table*}[t]
\centering
\caption{Detailed performance metrics of MDG across different benchmark splits}
\resizebox{\linewidth}{!}{
\begin{tabular}{l|cccccccc}
\toprule
\textbf{Benchmark} & \textbf{Score} & \textbf{Collision} & \textbf{TTC} & \textbf{Drivable area} & \textbf{Driving direction} & \textbf{Comfort} & \textbf{Ego progress} & \textbf{Speed limit} \\
\midrule
Val14 Non-Reactive          & 90.45 & 96.10 & 91.80 & 98.77 & 99.69 & 94.87 & 94.30 & 96.95   \\
Val14 Reactive              & 83.89 & 93.54 & 88.33 & 98.33 & 99.58 & 90.01 & 86.58 & 97.82 \\
Test14 Non-Reactive         & 90.16 & 95.93 & 91.46 & 98.78 & 99.69 & 94.51 & 94.08 & 96.85   \\
Test14 Reactive             & 83.21 & 93.23 & 87.70 & 98.15 & 99.59 & 89.34 & 86.08 & 97.85  \\
\bottomrule
\end{tabular}
}
\label{tab:metrics}
\end{table*}

\section{Experiment Details}
\subsection{Evaluation Metrics}
The evaluation metrics employed in the Waymo Sim Agents benchmark are detailed in \cite{montali2024waymo}. These metrics are designed to quantify the distributional divergence between ground-truth agent trajectories and simulated agent trajectories. The evaluation contains three key aspects: kinematic features, interactive features (e.g., time-to-collision (TTC), collision rate, distance to the nearest object), and map-based features (e.g., off-road and distance to road edge). A meta-composite realism score aggregates these components and serves as the primary evaluation metric.

For the nuPlan benchmark, we employ the closed-loop evaluation metrics defined in \cite{karnchanachari2024towards}. These include No At-Fault Collisions, Drivable Area Compliance, Making Progress, Driving Direction Compliance, TTC Within Bound, Progress Along Route Ratio, Speed Limit Compliance, and Comfort. The individual scores are weighted to produce the final Closed-Loop Score (CLS).

In addition to these benchmark-specific metrics, we report several supplementary measures to compare our model against state-of-the-art methods in open-loop settings. These metrics include:

\noindent\textbf{Collision Rate (CR).} This metric measures the average collision rate per scene (the number of colliding agents divided by the total number of agents), averaged by the total number of testing scenarios:
\begin{equation}
\text{CR} = \frac{1}{N_{\text{T}}} \sum_{N_{\text{T}}} \frac{1}{N_{\text{S}}} \frac{1}{N_{\text{A}}} \sum_{j=1}^{N_{\text{S}}}  \sum_{i=1}^{N_{\text{A}}} \mathbf{1}_{c}(s^j_i),
\end{equation}
where $N_{\text{T}}$ is the number of test scenarios, $N_{\text{S}}$ is the number of samples, $N_{\text{A}}$ is the total number of modeled agents in a scenario, and $s_i^j$ is the trajectory of agent $i$ of sample $j$. $\mathbf{1}_{c}(s^j_i)$ is an indicator function that equals 1 if the simulated trajectory results in a collision, and 0 otherwise.

\noindent\textbf{Off-road Rate (OR).} This metric quantifies the proportion of agents deviating off-road. It is calculated as the average off-road rate per scene (samples and agents), averaged by the total number of test scenarios:
\begin{equation}
\text{OR} = \frac{1}{N_{\text{T}}} \sum_{N_{\text{T}}} \frac{1}{N_{\text{S}}} \frac{1}{N_{\text{A}}} \sum_{j=1}^{N_{\text{S}}} \sum_{i=1}^{N_{\text{A}}} \mathbf{1}_{o}(s^j_i),
\end{equation}
where $\mathbf{1}_{o}(s^j_i)$ is an indicator function that equals 1 if the simulated trajectory veers off-road (e.g., outside the road boundary), and 0 otherwise. Note that we exclude those agents that are already off-road or labeled as pedestrians.

\noindent\textbf{Scene Average Displacement Error (SADE).} SADE computes the average Euclidean distance (L2 norm) between logged ground-truth trajectories and simulated trajectories across all agents and scenarios:
\begin{equation}
\text{SADE} = \frac{1}{N_{\text{T}}} \sum_{N_{\text{T}}} \frac{1}{N_{\text{S}}} \frac{1}{N_{\text{A}}} \sum_{j=1}^{N_{\text{S}}}  \sum_{i=1}^{N_{\text{A}}} \| s^j_i - s^{gt}_i \|_2,
\end{equation}
where $s^{gt}_i$ represents the ground-truth trajectory of an agent.

\noindent\textbf{Minimum SADE (minSADE).} This metric captures the minimum SADE (averaged over objects and minimum over samples), providing a measure of the best-case alignment with ground truth.
\begin{equation}
\text{minSADE} = \frac{1}{N_{\text{T}}} \sum_{N_{\text{T}}} \min_j \frac{1}{N_{\text{A}}} \sum_{i=1}^{N_{\text{A}}} \| s_i^j - s^{gt}_i \|_2.
\end{equation}

\noindent\textbf{Goal-reach Rate (GR).} This metric measures the average number of agents that successfully reach their assigned goals across all scenarios. An agent is considered to have reached its goal if the distance between its position and the goal position is less than 1 meter. The final value is obtained by averaging over the total number of scenarios.

\subsection{Baseline Methods for Open-loop Tasks}
\textbf{SMART} \cite{wu2024smart} is a state-of-the-art autoregressive next-token generation model for traffic agent simulation, achieving top performance on the Waymo Sim Agents benchmark. In our experiments, we use the tiny variant of SMART with 7M parameters. We adopt the open-source implementation provided in CAT-K \cite{zhang2024closed} and follow the default training and inference settings. To ensure a fair comparison, we do not perform closed-loop fine-tuning and only use the first-stage behavior cloning training.

\noindent\textbf{VBD} \cite{huang2024versatile} is a diffusion-based generative model for traffic agent simulation and ranked second in the 2024 Waymo Sim Agents Challenge. VBD performs joint trajectory diffusion, where all agents at a given diffusion step share the same noise level. We reproduce the model using the authors’ open-source implementation and evaluate it with five diffusion steps, generating the final predictions under this configuration.

\noindent\textbf{MTR} \cite{shi2022motion} is a high-performance motion prediction model achieving state-of-the-art results on the Waymo Motion Prediction benchmark. We adapt MTR for multi-agent simulation by using the same encoder as in our model and extending the decoder to produce predictions for all agents simultaneously. Predefined trajectory anchors serve as initial queries for each agent, allowing marginal predictions for all agents in a single forward pass without iterative querying.

\section{Additional Results}
\subsection{Ablation Studies}

\noindent\textbf{Influence of the Auxiliary Predictor.}
To examine the effect of the auxiliary trajectory predictor in the model, we conduct an ablation study by removing the predictor head from the model. As summarized in \cref{tab:predictor_influence}, excluding the predictor leads to a consistent decline in performance across both benchmarks. On the Waymo Sim Agents evaluation, the realism meta-metric drops from 0.7842 to 0.7682, while minADE increases from 1.3123\,m to 1.4117\,m, reflecting degraded simulation realism and motion accuracy. Similarly, on the nuPlan Val14 benchmark, CLS decreases from 89.75 to 86.90, indicating reduced planning performance. These results demonstrate that the auxiliary predictor enhances representation learning within the scene encoder by providing trajectory-level supervision.

\noindent\textbf{Influence of Action-to-State Conversion.}
The Transformer-based denoiser operates on a spatiotemporal representation of agent behaviors. To better capture motion dynamics, a differentiable kinematic function converts agent actions (e.g., acceleration and yaw rate) into corresponding physical states $(x, y, \theta, v)$, which are then encoded and denoised. We evaluate the impact of this action-to-state conversion on multi-agent open-loop trajectory prediction. All models are trained with five noise levels and evaluated using five-step temporal denoising to ensure consistent settings.
As shown in \cref{tab:physical}, integrating the conversion module substantially improves all evaluation metrics. Specifically, incorporating physical state transformation reduces collision and off-road rates by more than 40\% and decreases both SADE and minSADE by a significant margin. This improvement highlights the importance of embedding kinematic consistency within the denoiser: the conversion enables the model to reason over sequential dynamics rather than isolated action tokens, resulting in more physically coherent and accurate trajectory generation.

\noindent\textbf{Influence of Model Scale.}
We examine the impact of model scale on open-loop prediction performance by varying the hidden dimension of the Transformer while keeping the number of layers fixed. All models are trained for the same number of epochs, with batch sizes adjusted to fit GPU memory constraints. As shown in \cref{tab:scale}, enlarging the hidden dimension from 128 to 256 consistently improves performance across all metrics. However, further scaling to 512 yields no additional gains, indicating diminishing returns.
This saturation may stem partly from smaller batch sizes used at larger scales, but the primary factor is likely the limited training data, which restricts the model’s ability to fully leverage its increased capacity.

\noindent\textbf{Influence of Attention in Denoiser.}
We analyze the contribution of each attention module in the Transformer denoiser. As reported in \cref{tab:attn_ablation}, removing any of the three components consistently degrades prediction performance across all metrics, while the full model achieves the best overall results. These findings highlight the complementary roles of intra-agent, inter-agent, and agent-scene attention in the denoiser.

\begin{figure*}[ht]
    \centering
    \includegraphics[width=\linewidth]{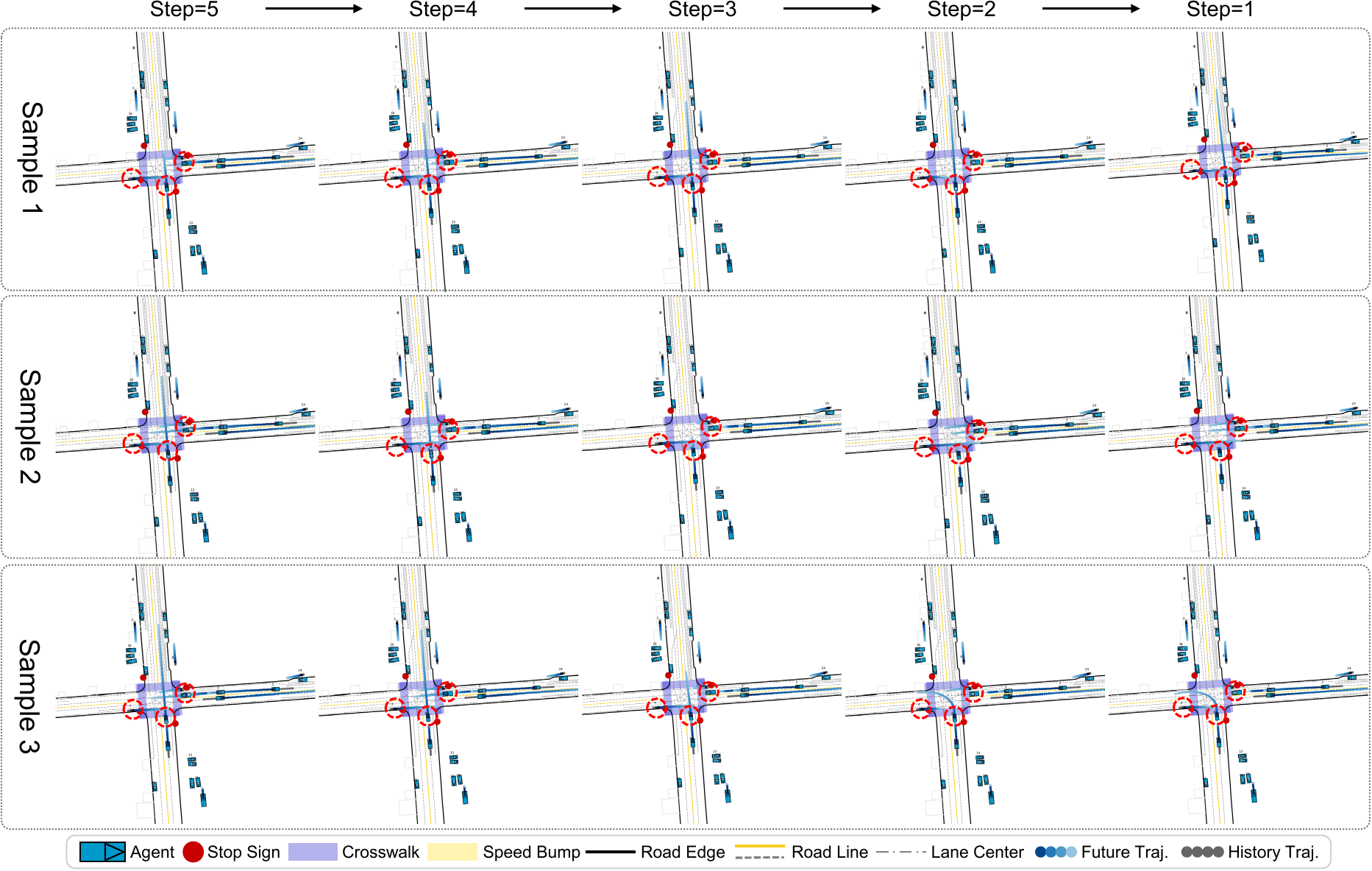}
    \caption{Visualization of the denoising process over time with five steps. Clean, denoised samples predicted by the MDG model at each step are shown. The process begins with similar predictions and gradually evolves to generate diverse future outcomes, emphasizing temporal behavior refinement and increased sample diversity across the time horizon.}
    \label{fig:dtime}
\end{figure*}

\begin{figure*}[ht]
    \centering
    \includegraphics[width=\linewidth]{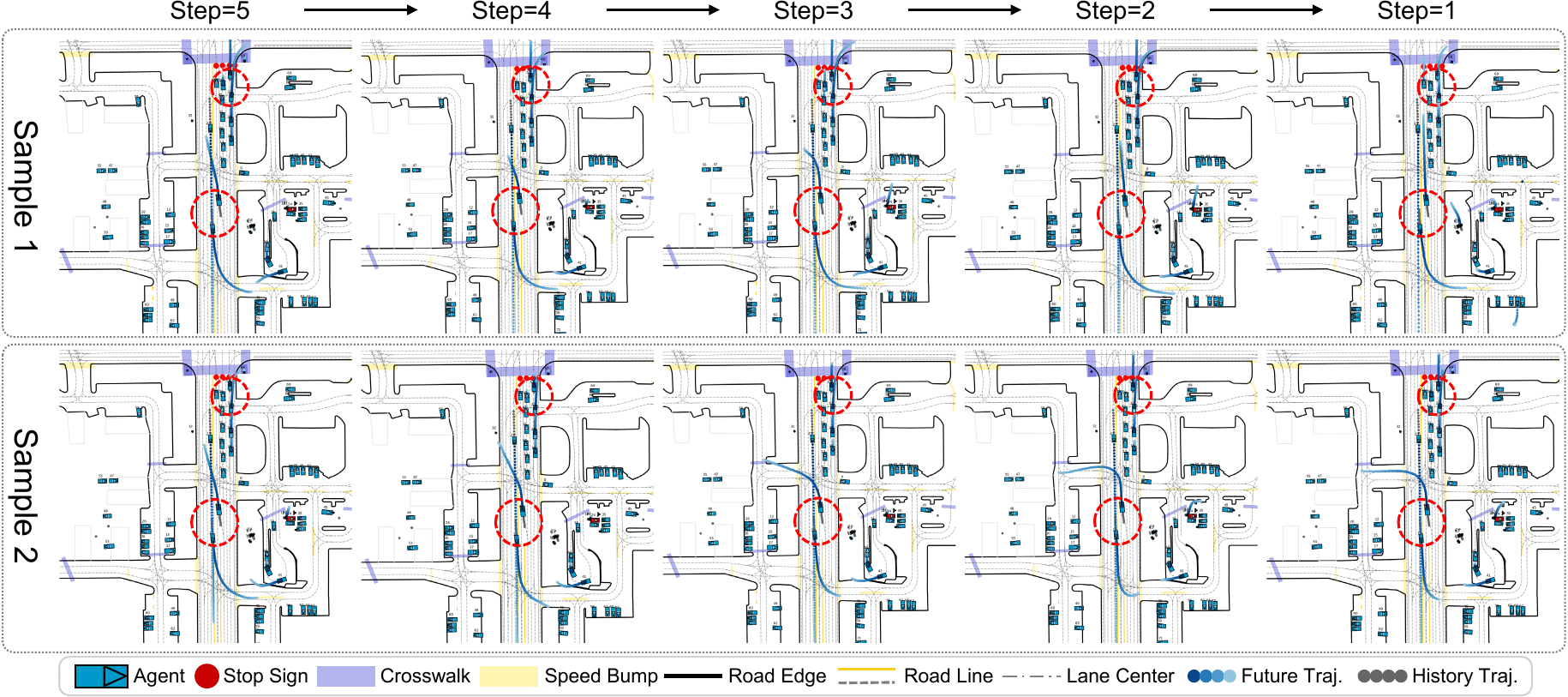}
    \caption{Visualization of the denoising process across agents over five steps. Clean, denoised samples predicted by the MDG model at each step are shown. Initial predictions for agents with high noise levels are suboptimal, but iterative refinement improves trajectory accuracy and enhances the diversity of behaviors.}
    \label{fig:dagents}
\end{figure*}

\begin{figure*}[ht]
    \centering
    \includegraphics[width=\linewidth]{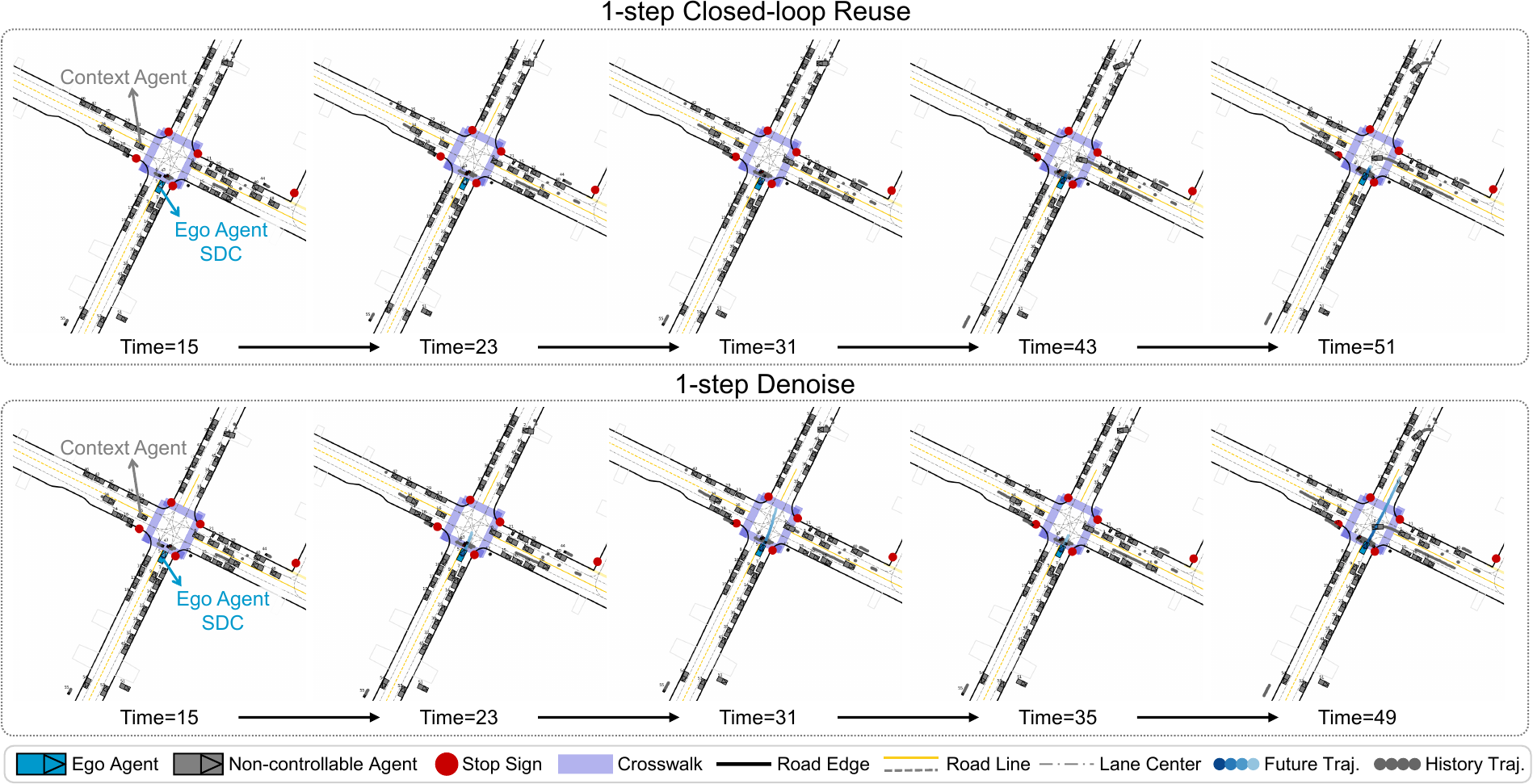}
    \caption{Comparison of a simple one-step denoising method and the closed-loop one-step reuse denoising method. The reuse method demonstrates improved temporal consistency and smoother planning trajectories for the ego agent.}
    \label{fig:reuse}
\end{figure*}

\begin{figure*}[ht]
    \centering
    \includegraphics[width=\linewidth]{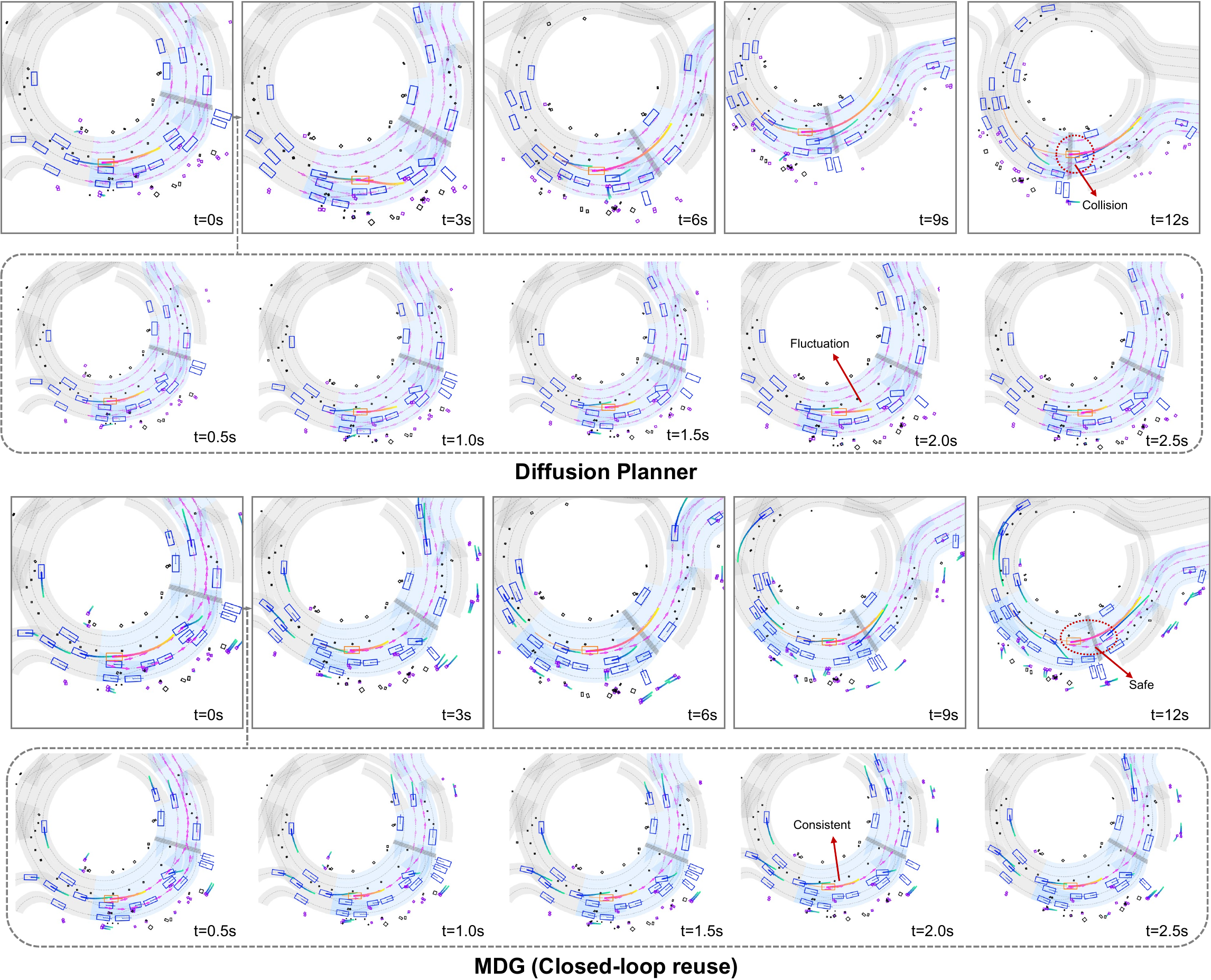}
    \caption{Comparison between MDG and Diffusion Planner in a nuPlan scenario. \textbf{Top:} The Diffusion Planner produces temporally inconsistent trajectories, leading to a collision. \textbf{Bottom:} MDG maintains consistent trajectories through a closed-loop denoising strategy and successfully completes the scenario without collisions.}
    \label{fig:plan}
\end{figure*}

\subsection{Visualization}
\textbf{Denoising Process.}  
We provide visualizations of the denoising process to illustrate its behavior. \cref{fig:dtime} depicts the temporal denoising process over five steps. Initially, the denoising results exhibit similarities across predictions. However, as the process progresses, subsequent denoising steps yield increasingly distinct outcomes for future predictions. Temporal denoising primarily focuses on the agents' dynamic behaviors, and increasing the number of denoising steps enables greater variation in intermediate actions. This, in turn, enhances the behavioral diversity across the entire time horizon.
In \cref{fig:dagents}, we demonstrate the denoising process along the agents over five steps. Unlike temporal denoising, agent-axis denoising emphasizes iterative trajectory-level refinement. Early denoising results are suboptimal for agents with high noise levels. However, as the denoising progresses, the trajectories become increasingly refined. This iterative process leads to a gradual determination of agents' behaviors, ultimately producing diverse and multi-modal joint agent trajectories.

\noindent \textbf{Closed-loop Reuse.}  
An example of the closed-loop reuse denoising method in the Waymo dataset is illustrated in \cref{fig:reuse}. In the case of a simple one-step denoising method, the ego agent's planning results exhibit significant variability across consecutive frames, even with small intervals, indicating poor temporal consistency. In contrast, the closed-loop one-step reuse method demonstrates improved temporal consistency, yielding smoother planning trajectories and better overall planning performance.

\noindent \textbf{Comparison on nuPlan Data.}
We compare MDG and the Diffusion Planner \cite{zheng2025diffusionbased} in a representative nuPlan scenario, as shown in \cref{fig:plan}. The Diffusion Planner generates trajectories that lack temporal coherence across planning steps, causing the ego vehicle to drift from a safe course and ultimately collide with surrounding traffic. In contrast, MDG preserves temporal consistency through its closed-loop denoising mechanism, enabling stable motion planning and safe scenario completion.

\subsection{Detailed Benchmark Results}
\cref{tab:metrics} presents the detailed metrics of the MDG (1-step closed-reuse) method on the nuPlan benchmarks. Performance on reactive settings is lower primarily due to the simplistic IDM used for reactive agents, which produces unrealistic behaviors and artificial gaps.

\end{document}